\documentclass[lettersize,journal]{IEEEtran}
\usepackage{amsmath,amsfonts}
\usepackage{algorithmic}
\usepackage{algorithm}
\usepackage{array}
\usepackage[caption=false,font=normalsize,labelfont=sf,textfont=sf]{subfig}
\usepackage{textcomp}
\usepackage{stfloats}
\usepackage{url}
\usepackage{verbatim}
\usepackage{graphicx}
\usepackage{cite}
\usepackage{multirow}
\usepackage[normalem]{ulem}
\usepackage{booktabs}
\usepackage{balance}

\usepackage{color}
\usepackage[table]{xcolor}
\usepackage{soul}
\begin{document}

\title{PPGF: Probability Pattern-Guided Time Series Forecasting}

\author{Yanru Sun, Zongxia Xie, Haoyu Xing, Hualong Yu, Qinghua Hu,~\IEEEmembership{Senior Member,~IEEE,}
\thanks{This paper was produced by the IEEE Publication Technology Group. They are in Piscataway, NJ.}
\thanks{This work was supported in part by the National Natural Science Foundation of China under Grant 62376194, Grant 61925602, and Grant U23B2049. (Corresponding authors: Zongxia Xie.)}
\thanks{Yanru Sun, Zongxia Xie, Haoyu Xing, Hualong Yu, and Qinghua Hu are with the College of Intelligence and Computing and the Tianjin Key Laboratory of Machine Learning, Tianjin University, Tianjin 300350, China (e-mail: yanrusun@tju.edu.cn; caddiexie@hotmail.com; hyxing@tju.edu.cn; yhl2000@tju.edu.cn; huqinghua@tju.edu.cn).}}

\markboth{IEEE TRANSACTIONS ON NEURAL NETWORKS AND LEARNING SYSTEMS}%
{Shell \MakeLowercase{\textit{et al.}}: A Sample Article Using IEEEtran.cls for IEEE Journals}


\maketitle

\begin{abstract}
Time series forecasting (TSF) is an essential branch of machine learning with various applications. Most methods for TSF focus on constructing different networks to extract better information and improve performance. 
However, practical application data contain different internal mechanisms, resulting in a mixture of multiple patterns. That is, the model's ability to fit different patterns is different and generates different errors. 
In order to solve this problem, we propose an end-to-end framework, namely probability pattern-guided time series forecasting (PPGF). 
PPGF reformulates the TSF problem as a forecasting task guided by probabilistic pattern classification. 
Firstly, we propose the grouping strategy to approach forecasting problems as classification and alleviate the impact of data imbalance on classification. Secondly, we predict in the corresponding class interval to guarantee the consistency of classification and forecasting. In addition, True Class Probability (TCP) is introduced to pay more attention to the difficult samples to improve the classification accuracy. Detailedly, PPGF classifies the different patterns to determine which one the target value may belong to and estimates it accurately in the corresponding interval. 
To demonstrate the effectiveness of the proposed framework, we conduct extensive experiments on real-world datasets, and PPGF achieves significant performance improvements over several baseline methods. Furthermore, the effectiveness of TCP and the necessity of consistency between classification and forecasting are proved in the experiments. All data and codes are available online: \url{https://github.com/syrGitHub/PPGF}.
\end{abstract}

\begin{IEEEkeywords}
Time Series Forecasting, Probability Pattern Classification, Simultaneous Classification and Forecasting.
\end{IEEEkeywords}

\section{Introduction}
\IEEEPARstart{T}{ime} series forecasting (TSF), based on observed historical data to predict numerical value fluctuations over time, has attracted growing attention in recent years. It plays a crucial role in many real-world applications, including traffic flow forecasting \cite{long2024unveiling, kong2024spatio}, air quality supervision \cite{patel2022accurate, yu2025mgsfformer}, weather \cite{sun2022solar, bi2023accurate, wu2023interpretable, lam2023learning}, and others \cite{akhter2022hour,wang2022hour, wang2023accurate, liu2023cross}. 

At present, Deep Neural Network (DNN)-based techniques have been developed for TSF tasks \cite{liu2024deep, qiu2024tfb}.
As an essential component of these models, Recurrent Neural Network (RNN) \cite{elman1990finding} is an indispensable factor for these note-worthy improvements. Subsequently, the variants of RNN, Long Short Term Memory (LSTM) \cite{graves2012long} and Gated Recurrent Unit (GRU) \cite{chung2014empirical}, have significantly improved the state-of-the-art performance because they can effectively capture both long- and short-term dependencies. 
However, RNN can only process one-time step at a time and result in an exceptionally computationally intensive.
Gradually, in order to deal with high dimensional data efficiently, Convolutional Neural Network (CNN)-based methods have been developed \cite{wutimesnet, luo2024moderntcn}.
Graph Neural Network (GNN) \cite{shao2022pre,li2023dynamic} has been used to capture specific node patterns and automatically infer the dependencies between sequences.
In addition, Transformer \cite{zhou2021informer, wu2021autoformer, nie2022time, liu2024itransformer} has been proposed as a new architecture that utilizes attention mechanisms to process sequential data.

\begin{figure}[!t]
		\centering
	\begin{minipage}[t]{0.9\linewidth}
		\centering
		\includegraphics[width=\textwidth]{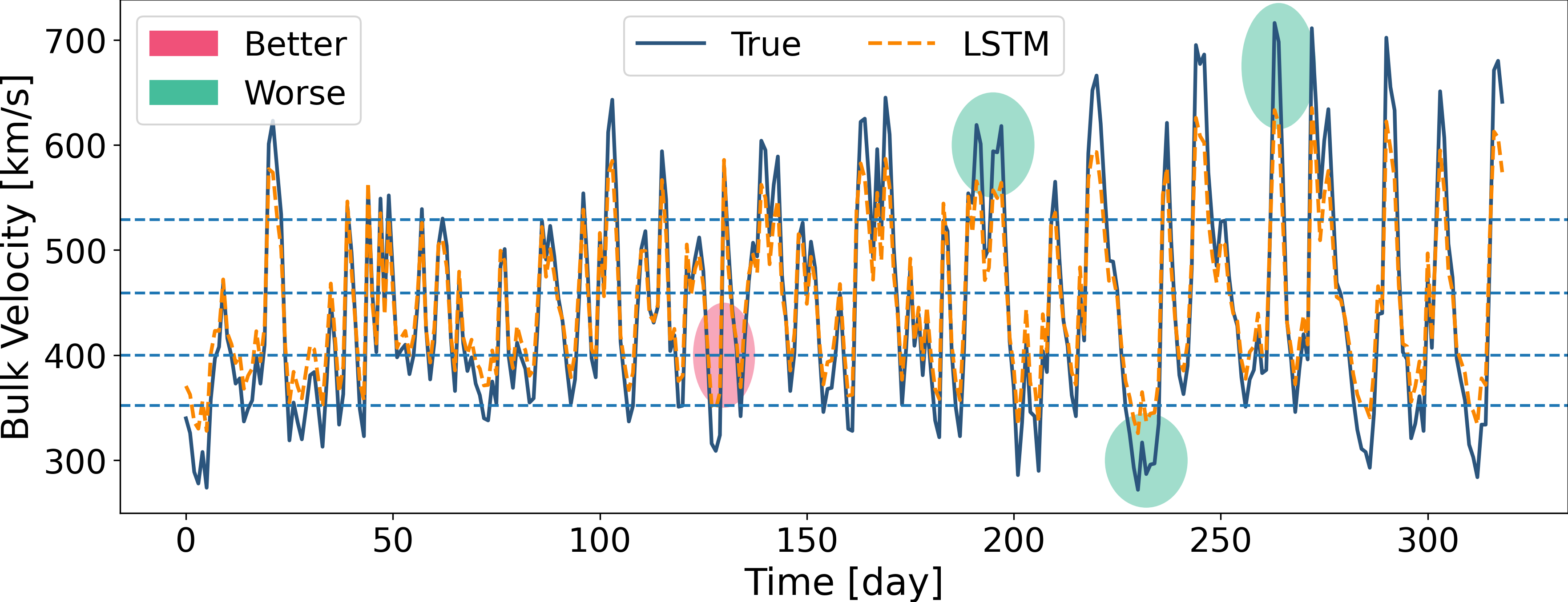}
		\centerline{(a)}
		\vspace{0.3mm} 
	\end{minipage}
	\begin{minipage}[t]{0.48\linewidth}
		\centering
		\includegraphics[width=\textwidth]{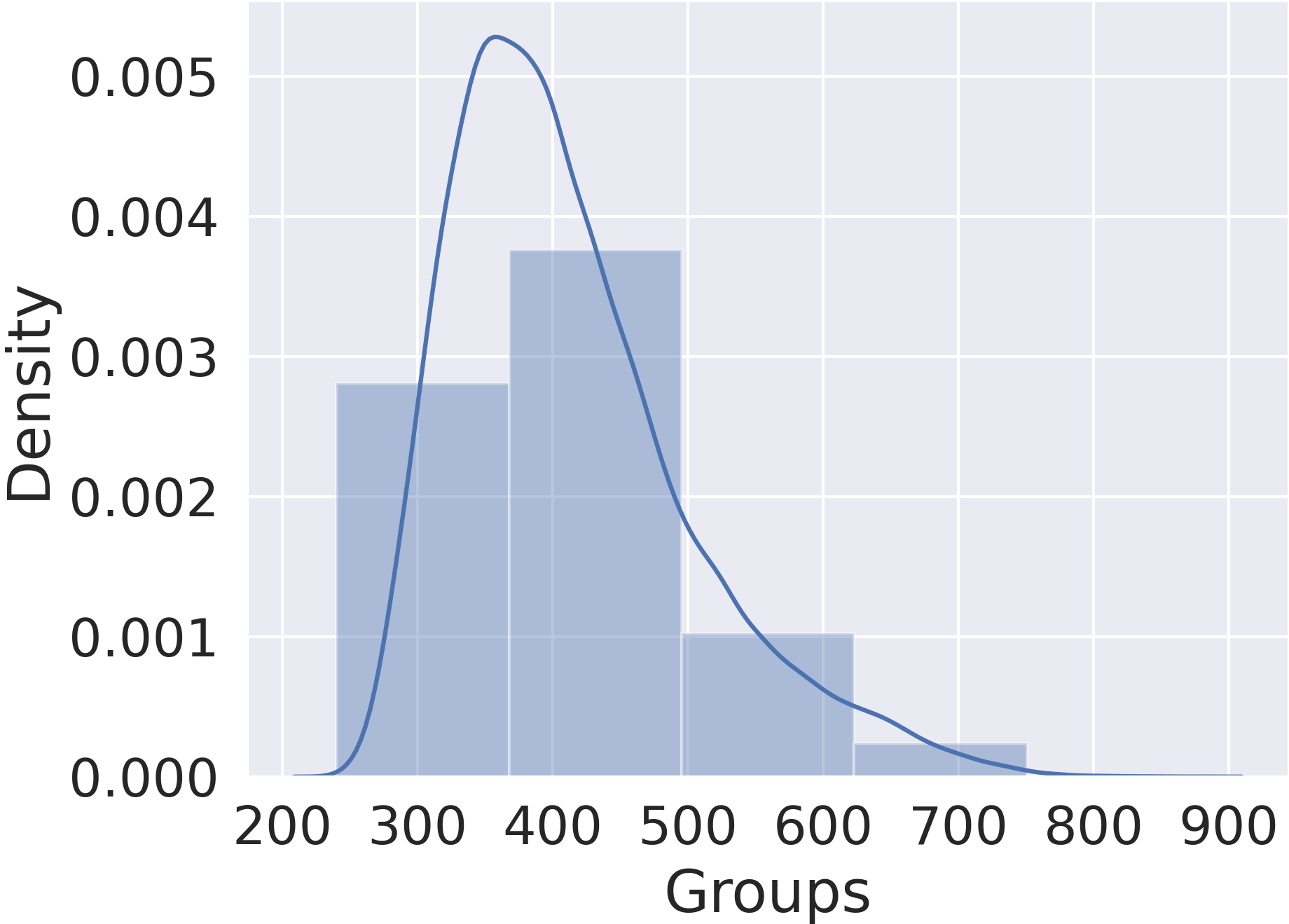}
		\centerline{(b)}
	\end{minipage}\hspace{1mm}
	\begin{minipage}[t]{0.48\linewidth}
		\centering
		\includegraphics[width=\textwidth]{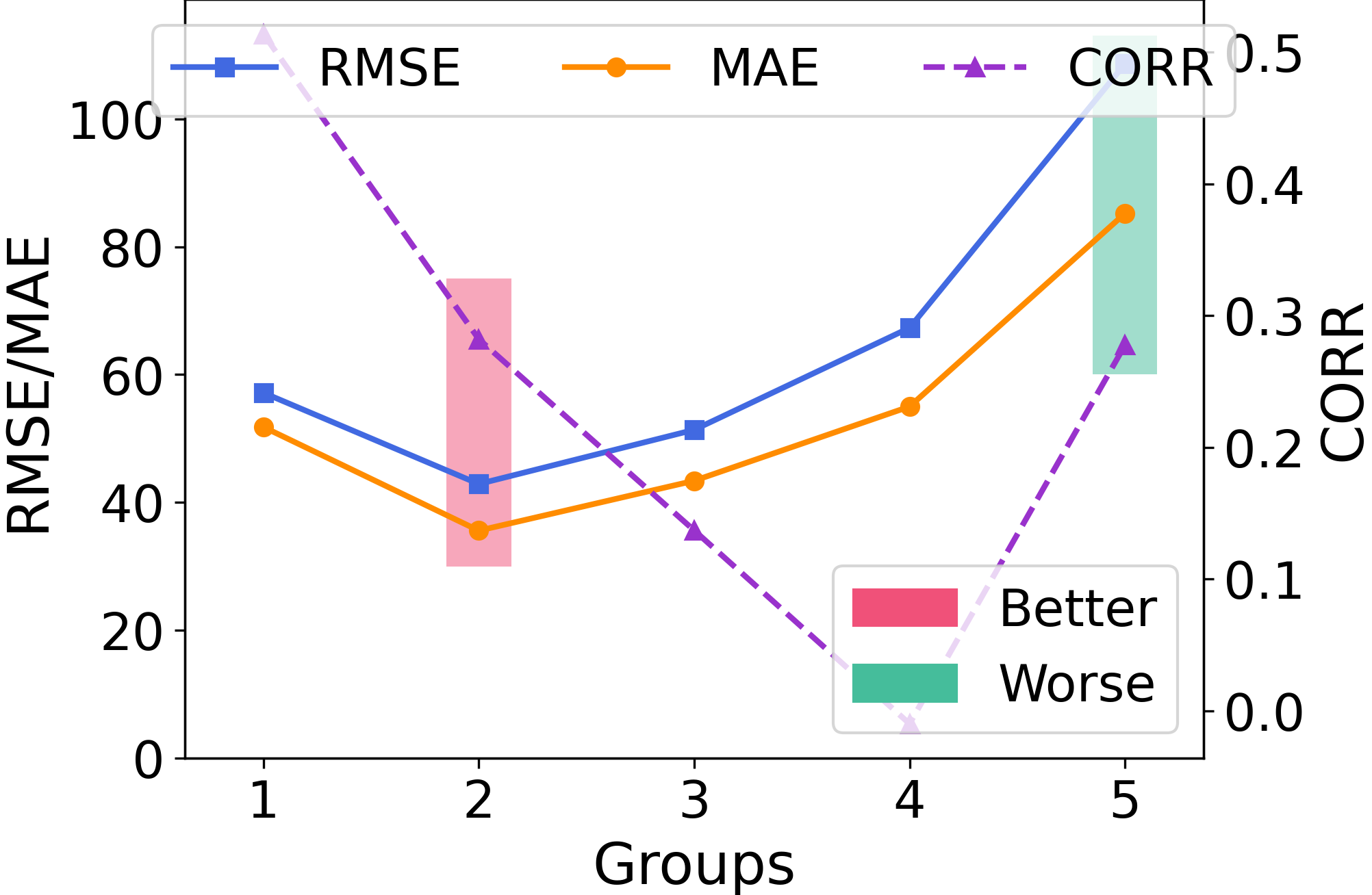}
		\centerline{(c)}
	\end{minipage}
	\caption{Relationship between model accuracy and different patterns. We divided the data into five groups representing five patterns. (a) LSTM has different abilities to fit data in various intervals. (b) Time series data are imbalanced and have a long-tailed distribution. (c) The performance of LSTM is disparate for various patterns. The prediction performance is worse for the patterns with few samples, while it is better for the patterns with more samples.}
	\label{fig:Relation}
\end{figure}

The traditional methods consider all the time series samples as a whole and focus on constructing different networks to extract better features to improve performance. However, real-world applications often contain different internal mechanisms, resulting in a mixture of multiple patterns \cite{li2022extreme, liu2024adaptive}. For example, the slow-speed, steady solar wind caused by coronal holes and the high-speed solar wind caused by coronal mass ejections belong to different patterns due to different generation causes \cite{sun2022accurate}. The direct way is to introduce classification to classify different patterns. As illustrated in Figure~\ref{fig:Relation}, we divide the solar wind speed into five groups to represent different patterns according to their magnitude, and a standard LSTM is trained to predict solar wind speed. We can see that 1) Figure~\ref{fig:Relation}~(a): the model has different abilities to fit data in various patterns; 2) Figure~\ref{fig:Relation}~(b): the time series data have a long-tailed distribution after grouping; 3) Figure~\ref{fig:Relation}~(c): the model has different performances on different patterns, and patterns with more samples performed better. Therefore, there are three challenges when considering the different patterns in TSF, which can be described as follows: 

\textbf{Challenge 1: Pattern Grouping.} How to define a general pattern grouping method is an important question. Abraham et al.\cite{abraham2009semi} consider the zeros and the non-zero values in the data. Tan et al. \cite{tan2018geomagnetic} segment the data set into geomagnetic storm occurrence and non-occurrence based on the $Kp$ geomagnetic index. Wilson et al.\cite{wilson2022beyond} divide the data into extreme and non-extreme events according to a variable threshold $U$. In addition, DXtreMM \cite{abilasha2022deep} sets extreme thresholds for all datasets as $2^{nd}$ quantile for left extreme and $98^{th}$ quantile for right extreme. However, most of these methods rely on domain knowledge and cannot be generalized to different data sets. In addition, most of them do not take into account the effects of data imbalance on classification.

\textbf{Challenge 2: Classification Precision.} Through pattern grouping, we introduce the classification to mine the patterns in the data, and its accuracy is critical. If the classified pattern is incorrect, the predicted value will be modeled differently, which causes a significant error. So, it is essential to estimate whether a model makes an error and calibrate the classification results.

When we consider introducing the classification into forecasting tasks, it generates another challenge.

\textbf{Challenge 3: Consistency of Classification and Forecasting.} Some algorithms also consider the multiple patterns contained in time series and decompose TSF into classification and forecasting problems. At first, the two-step approaches are proposed, which train the classifier and the predictors separately \cite{tan2018geomagnetic,abilasha2022deep}. However, these methods require a lot of time and space to train different predictors, and the classifier's performance dramatically affects the model's performance. Recently, some researchers have proposed end-to-end methods that optimize the joint objective function that considers both classification errors and prediction errors\cite{yu2021group,jiang2021multi,ben2023simultaneous}. However, these methods have no constraints between classification and forecasting so the prediction and category may belong to a different interval.

To address these challenges, we propose an end-to-end framework, called \textbf{P}robability \textbf{P}attern-\textbf{G}uided time series \textbf{F}orecasting (PPGF).
For \textbf{pattern grouping}, \textit{Grouping Strategy} is introducing to divide the whole range into multiple groups and avoid the impact of data imbalance on classification.
For \textbf{classification precision}, we construct a novel structure, \textit{Probability Pattern Classifier}, which introduces True Class Probability (TCP) \cite{corbiere2019addressing} to estimate whether a model makes an error and produce the features' reliability confidence to calibrate the feature. Then the classifier adopts the calibrated feature to achieve the classification task and obtain the category $k$.
For \textbf{consistency}, \textit{Relative Prediction Strategy} is designed to produce the relative score $\Delta Y$ to the lower bound of category $k$. Then, we propose \textit{Probability Pattern-guided Forecasting} to add the consistency constraint between classification and forecasting, guaranteeing the prediction value is in the corresponding class interval.
The final prediction is obtained according to $k$ and $\Delta Y$.
PPGF incorporates three loss functions, namely confidence loss ($\mathcal L^{conf}$), cross-entropy loss ($\mathcal L^{cls}$), and mean square error (MSE) loss ($\mathcal L^{reg}$), which balances the impact of true label probability, classification, and forecasting.

To summarize, the main contributions of this work are as follows:
\begin{enumerate}
    \item We introduce a probability pattern-guided time series forecasting (PPGF) framework that aligns classification and forecasting processes, ensuring predictions fall within the corresponding classification intervals.
    \item This article proposes a grouping strategy to mitigate the effects of data imbalance, which does not require expert knowledge and can be flexibly applied to various datasets.
    \item The proposed PPGF framework uses TCP approximation to estimate the confidence of the classification and calibrate the features accordingly, which can improve pattern classification accuracy effectively.
    \item Extensive experiments on real-world time series datasets demonstrate that our framework significantly outperforms state-of-the-art methods, while qualitative results validate its effectiveness and interpretability.
\end{enumerate}

\section{RELATED WORK}
\label{sec:Related Work}
\subsection{Time Series Forecasting}
The past few years have witnessed the rapid development of TSF. Time series forecasting has many challenges, such as capturing long-term dependencies, noise sensitivity, computational efficiency, and so on \cite{nie2022time,eldeletslanet}. The prevalent methods in TSF can be categorized into three main groups: Classical methods, Transformer-based methods, and Frequency-domain methods.

\textbf{Classical methods.} Bai et al. \cite{bai2018empirical} propose a Temporal Convolutional Network (TCN) architecture combined with CNN, which achieves better performance than RNN while avoiding the common drawbacks of recursive models, such as the gradient explosion/disappearance problem.
Dual Self-Attention Network (DSANet) \cite{huang2019dsanet} is proposed for multi-variable TSF, which uses two parallel convolution components to capture a complex mixture of global and local temporal patterns.
Temporal Pattern Attention mechanism (TPA) is proposed in \cite{shih2019temporal}, which extracts the fixed-length temporal patterns in the input information by using CNN filters. TPA determines the weights of each temporal pattern using a scoring function, and obtains the final output according to the magnitude of the weights.
Additionally, the Multiscale recurrent network (MRN) \cite{guo2023multivariate} seamlessly integrates multiscale analysis into deep learning frameworks to build scale-aware recurrent networks.
A novel sequence-to-sequence-based deep learning model utilizing a time series decomposition strategy for multistep load forecasting is proposed in \cite{lu2024novel}. This model can provide better insights for optimizing energy resource allocation and assisting in the decision-making process.
Furthermore, Bai et al. \cite{bai2020adaptive} propose Adaptive Graph Convolutional Recurrent Network (AGCRN) combined with GNN to capture the fine-grained spatio-temporal correlations of traffic sequences automatically.

\textbf{Transformer-based methods.} The emergence of Transformer architectures has revolutionized the ability to capture long-term dependencies in time series data. Li et al. \cite{li2019enhancing} propose the transformer-based architecture to process time series data capturing the long-term dependencies. PatchTST \cite{nie2022time} segments the time series into subseries-level patches and adopts the channel-independence to improve the long-term forecasting accuracy. iTransformer \cite{liu2024itransformer} embeds the time points into variate tokens and applies the feed-forward network for each variate token to empower the Transformer family with promoted performance. However, DLinear \cite{zeng2023transformers} adopts a single linear layer to extract the dominant periodicity from the temporal domain and surpasses a range of advanced complex networks.

\textbf{Frequency-domain methods.} On the other hand, many methods have been proposed for TSF from the frequency domain, which are orthogonal to our methods. FITS \cite{xu2023fits} addresses time series forecasting through interpolation in the complex frequency domain, while FreDF \cite{wang2024fredf} avoids the complexity of label autocorrelation in the time domain by focusing on forecasting in the frequency domain.

Despite the progress in these areas, existing methods often treat time series data as single-pattern data, neglecting the diverse intrinsic mechanisms inherent in time series. This oversight results in varying learning capabilities for distinct patterns, ultimately leading to less accurate predictions. In contrast, we first mine the different patterns of the data and model them by predictor to enhance forecasting performance.

\subsection{Simultaneous Classification and Forecasting}
Unlike traditional single tasks, associative tasks can improve performance by sharing information and complementing each other. Specifically, the interaction of unrelated parts between tasks helps to escape local minima. In addition, the relevant parts between tasks facilitate learning generic feature representations at the bottom shared layer.

Combining classification with forecasting has demonstrated a significant potential in time series forecasting.
There are two main types. The first type is the two-step method. Tan et al. \cite{tan2018geomagnetic} propose the method to predict the $Kp$ geomagnetic index, which first predicts the outputs are storm or non-storm, and then uses two separate submodels to forecast the $Kp$ with and without a geomagnetic storm.
DXtreMM \cite{abilasha2022deep} improves the extreme events performance by composing a classifier and many forecasting units. The classifier predicts the possibility of occurrence of an extreme event given a time segment, and the forecasting module predicts the exact value.
However, these methods consume a lot of time and computing resources due to the training of multiple sub-models.

Furthermore, many strategies propose the end-to-end combination of classification and forecasting tasks to improve extreme events performance. 
Abraham et al. \cite{abraham2010integrated} present an integrated framework that simultaneously performs classification and forecasting to accurately predict future values of a zero-inflated time series.
In addition, DEMM \cite{wilson2022beyond} divides the data into three categories according to the magnitude and combines the hurdle model with the extreme value theory to simulate the distribution of different events while accurately making point prediction and distribution prediction.

However, these approaches do not explore the constrained relationship between the two tasks, i.e., even if the classification is correct to class $k_i$, the predicted result is likely to be class $k_j$. To address the limitation of the co-existence of two tasks, this paper presents a novel approach that introduces a guiding relationship between them.\\ 

\section{Method}
\label{sec:Methodology}
\subsection{Problem Description and Grouping Strategy}
\label{sec:problem}
\textbf{Problem Description.} Let $A = \{a_1,\ldots,a_t\}\in \mathbb{R}^{t\times D}$ be a multi-variate time series with channel $D \geq 1$. The data is split into (input, output) pairs using a sliding window approach. Formally, the input matrix of the common TSF method is $X = \{x_i\}_{i=1}^N$ with $N$ samples, where $x_i\in \mathbb{R}^{L\times D}$, $L$ is the length of the look-back window. The goal of TSF is to predict horizon series: $Y = \{y_i\}_{i=1}^N$, where $y_i\in \mathbb{R}^{T}$, $T$ is the length of the horizon.

\begin{figure}[!t]
		\centering
	\begin{minipage}[t]{0.48\linewidth}
		\centering
		\includegraphics[width=\textwidth]{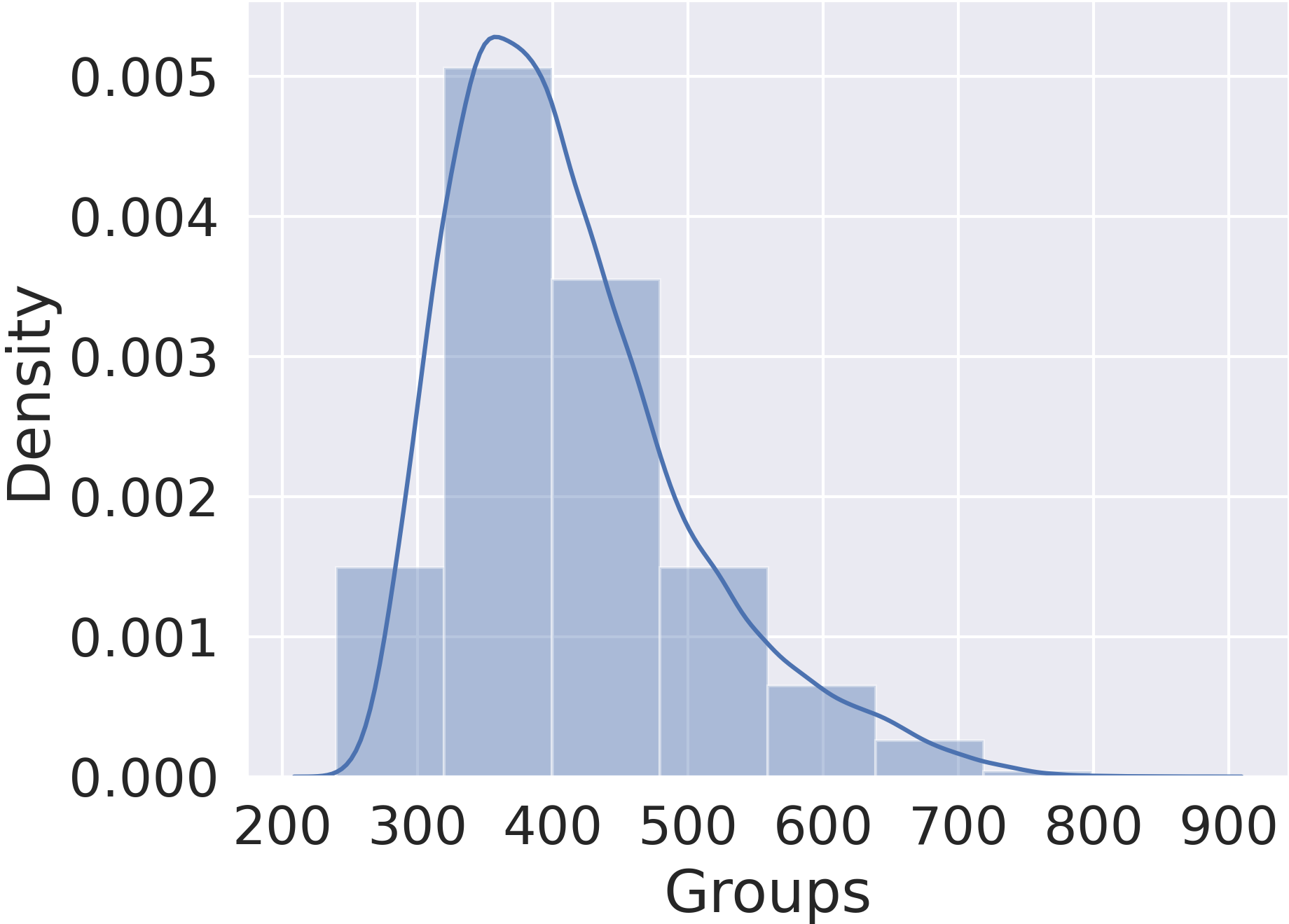}
		\centerline{(a)}
	\end{minipage}\hspace{1mm}
	\begin{minipage}[t]{0.48\linewidth}
		\centering
		\includegraphics[width=\textwidth]{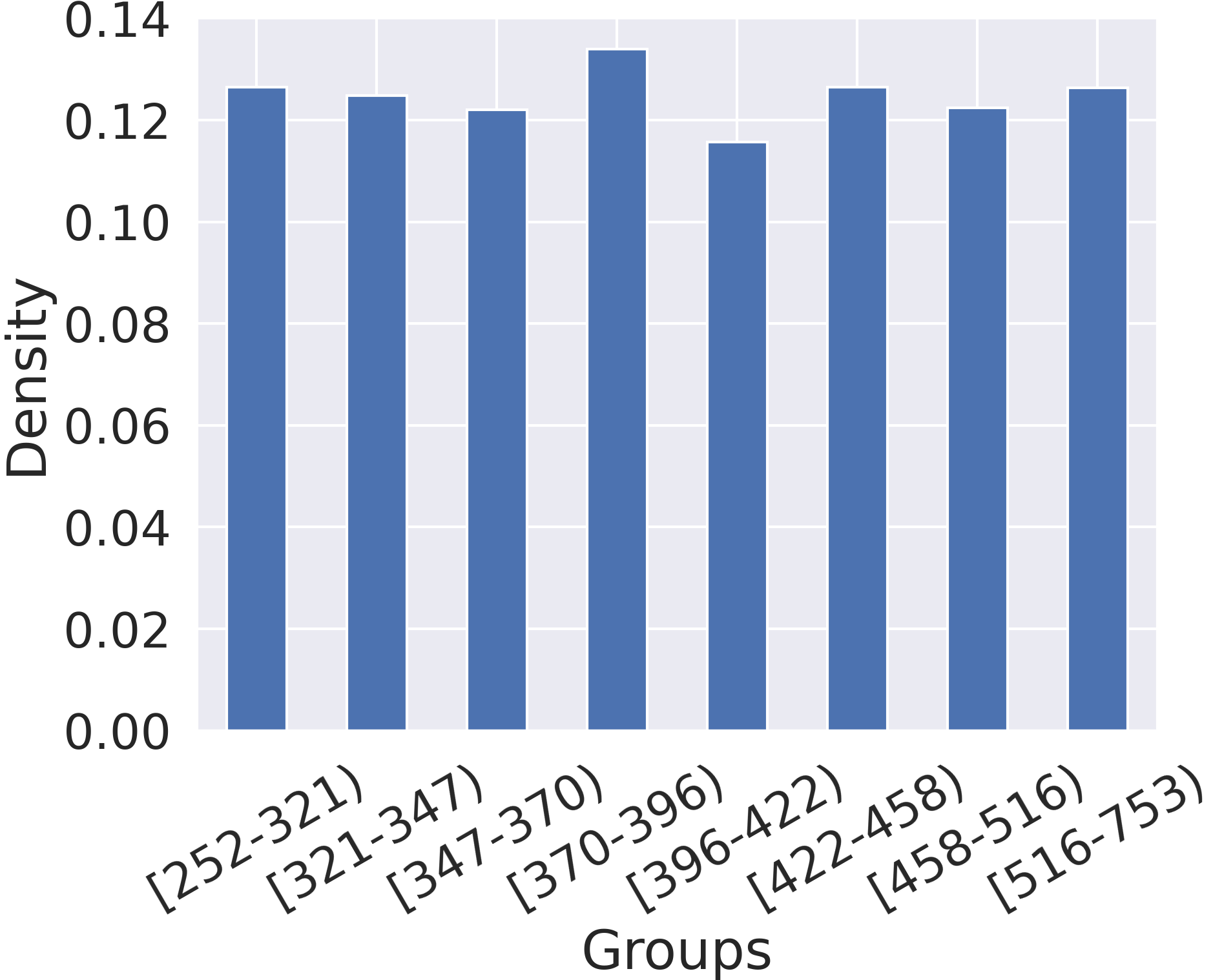}
		\centerline{(b)}
	\end{minipage}
	\caption{The distribution of the solar wind in the training set under different partition strategies. (a) Uniform partition. We can observe a large variety of frequencies among different groups. (b) The proposed grouping strategy is in Equation \ref{equ:group}. The data belonging to each group are balanced.}
	\label{fig:division strategy}
\end{figure}

\textbf{Grouping Strategy.} A common grouping strategy is the "equal width" strategy, which simply divides the whole interval into multiple equal-width groups. However, this method leads to serious data imbalance in each group, as shown in Figure~\ref{fig:division strategy} (a). 

To enhance the implementation of pattern grouping, we propose a new strategy that transforms complex temporal relationships into multiple pattern groups.
Specifically, we divide $A$ into $K$ non-overlapping intervals (namely "groups") to represent different patterns.
First, we collect the list of continuous values $A = \{a_1,\ldots,a_t\}$, then we arrange them in ascending order to get $\tilde{A} = \{\tilde{a_1},\ldots,\tilde{a_t}\}$. Given the group number $K$, the partitioning algorithm gives the boundary of each interval $\mathcal{I}_k = (\rho_k^{left}, \rho_k^{right})$ as:
\begin{align}
&\rho_k^{left}=\tilde{A} (\lfloor (t-1)\times \frac{(k-1)}{K} \rfloor),\\
&\rho_k^{right}=\tilde{A} (\lfloor (t-1)\times \frac{k}{K} \rfloor), \forall k=1,2,\ldots,K.
\label{equ:group}
\end{align}
where $t$ is the length of $A$ and we use $\tilde{A}(k)$ to represent the $k$-th element of $A$.

As shown in Figure~\ref{fig:division strategy} (b), the grouping method we propose has two advantages. First, the number of objects in each interval is approximately equal, which alleviates the impact of data imbalance on the results. Moreover, this method is domain-independent and does not rely on expert knowledge.

We can then obtain the target for the relative prediction strategy, denoted as $\Delta Y=\{\Delta y_i\}_{i=1}^N$, which represents the lower bound offset of $y$ with respect to each class $k$, where $\Delta y_i \in \mathbb{R}^{T}$.
Consequently, the new structure of the dataset is $D = \{x_i, y_i, \Delta y_i, k_i\}_{i=1}^N$, consisting of N samples.

\begin{figure*}[!t]
    \centering
    \includegraphics[width=0.8\linewidth]{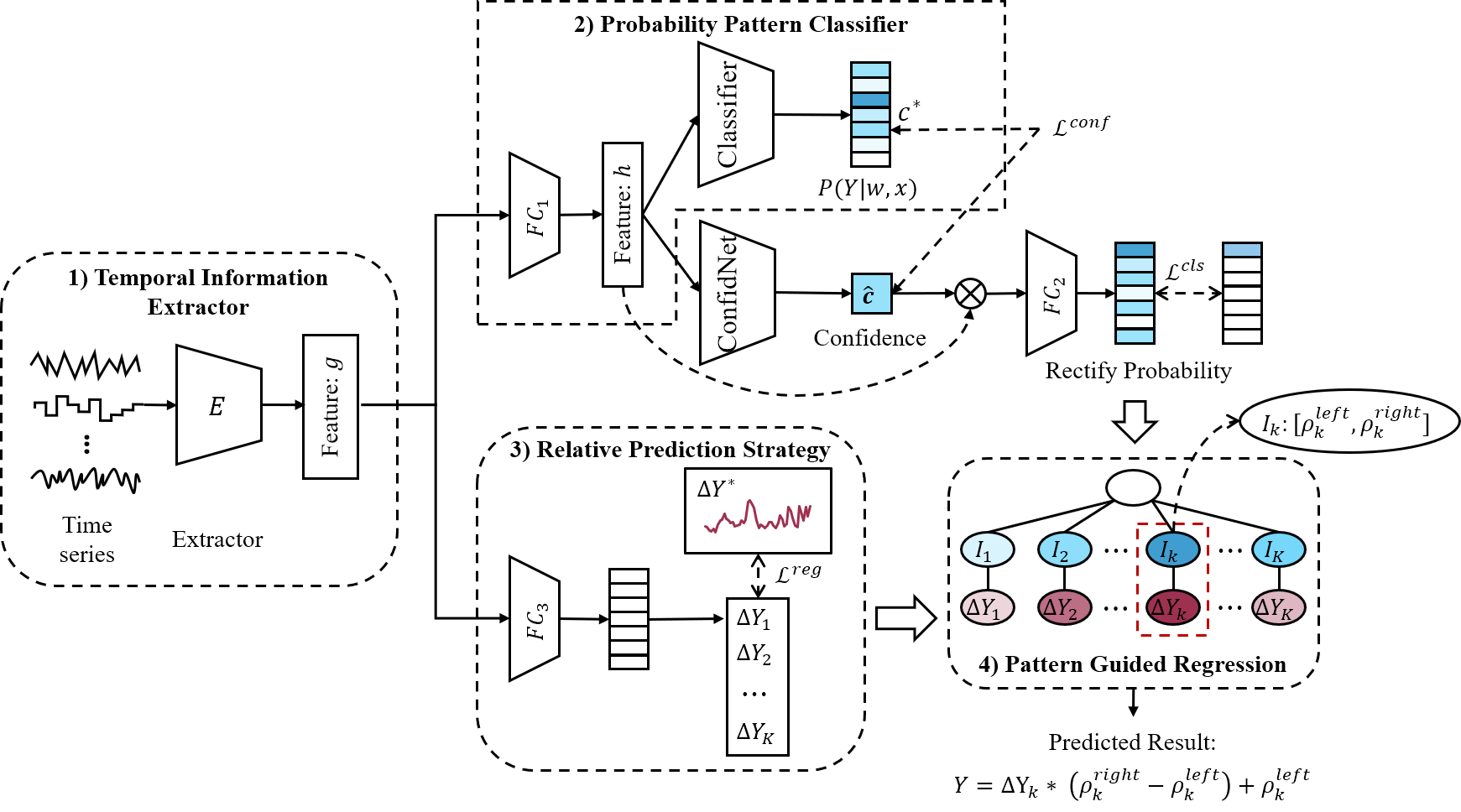}
    \caption{Pipeline of our proposed Probability Pattern-Guided Forecasting method. Firstly, Temporal Information Extractor captures general temporal features at the shallow layer of the network. Then, these features are fed into the Probability Pattern Classifier and relative prediction strategy, respectively, to learn discriminant features and more specific forecasting features. Finally, the probability Pattern-Guided Forecasting strategy is used to get the final prediction results.}
    \label{fig:Model}
\end{figure*}

\subsection{Temporal Information Extractor}
\label{sec:Temporal Information}
Firstly, we selected the combination of 1-D convolution, Transformer, and Gated Residual Network (GRN) as the temporal information extractor to capture the general features:

\begin{itemize}
\item We adopt the 1-D convolution to capture local dependencies of the input, allowing for rapid feature extraction without extensive computational resources \cite{dempster2020rocket}. Its ability to identify localized patterns makes it particularly advantageous in time series forecasting.
\item Following the convolutional layer, we employ a Transformer encoder to model long-term dependencies, which are critical for accurate time series forecasting \cite{zhou2021informer}. The combination of convolution and self-attention enhances the model’s ability to learn relationships across varying time scales. 
\item To further refine the model, we incorporate a GRN . Its gating mechanism provides adaptive depth and complexity, allowing the model to dynamically adjust to diverse datasets and scenarios. This adaptability is crucial for handling the heterogeneity often observed in time series data. 
\end{itemize}

By applying the temporal information extractor in the shallow layer of the network, we capture the general temporal information of the time series: $g\in \mathbb{R}^{B\times o}$, where $B$ indicates batch size and $o$ denotes output dimension. Compared to Temporal Convolutional Networks (TCN), which primarily rely on stacked convolutional layers with fixed receptive fields, our approach leverages the flexibility of self-attention in Transformers to dynamically model dependencies across varying time scales. This adaptability, combined with the gating mechanism in GRN, allows our model to better capture both local variations and complex long-term patterns inherent in diverse time series data \cite{bai2018empirical}.

\subsection{Probability Pattern Classifier}
\label{sec:Classification Information}
For time series data, it contains various patterns, and accurate classification is crucial for subsequent relative prediction. Therefore, it is essential for pattern classification to be aware of when an erroneous prediction has been made. In classification tasks, the confidence estimation baseline widely used by neural networks is to take the probability value of the prediction class given by the softmax layer output, namely the Maximum Class Probability (MCP):
\begin{align}
\mathcal L^{cls} = -\sum_{k=1}^{K} k \log p_k,
\end{align}
where $k$ is the class label. $\mathcal{L}^{cls}$ is the cross-entropy loss function. However, it is worth mentioning that MCP leads to high confidence values, especially for erroneous predictions making it hard to distinguish errors and correct predictions.

Based on the above motivation, we propose employing TCP \cite{corbiere2019addressing} to achieve more reliable classification confidence. TCP leverages the Softmax output probability associated with the true label as a measure of confidence. When the confidence of a classification is low, it indicates uncertainty, prompting us to potentially adjust the classification result.
Formally, given an input time series $x$, the network produces a probabilistic predictive distribution $P$ for each class, and TCP can be defined as follows:
\begin{align}
c^* = P(Y = k^*|w,x),
\end{align}
where $k^*$ is the real label and $w$ are the parameters of the network. Both TCP and MCP represent the largest Softmax outputs for samples that the model correctly classifies, serving as good indicators of classification confidence. However, when the model misclassifies an example, MCP can produce high confidence values for both incorrect and correct predictions, making it challenging to differentiate between them. In contrast, the probability associated with the true label $k^*$ is often closer to a lower value, indicating that the model is likely making an incorrect prediction \cite{corbiere2019addressing}.

Nevertheless, the true label $k^*$ of the output is unavailable when estimating the confidence of the test sample. To address this, we introduce a confidence neural network, termed ConfidNet, with parameters $\theta$. This network outputs a confidence prediction $\hat{c}(h, \theta)$. During training, we optimize $\theta$ using the training samples to ensure $\hat{c}(h, \theta)$ approximates $c^*(h, k^*)$. To achieve a regression score between 0 and 1, we train the confidence neural network using $\ell_2$ loss:
\begin{align}
\mathcal L^{conf} = (\hat{c}(h, \theta)-c^*(h, k^*))^2.
\end{align}
Then the true label probability can be approximated with the classifier and confidence estimation network.

\begin{figure}[!t]
    \centering
    \includegraphics[width=0.8\linewidth]{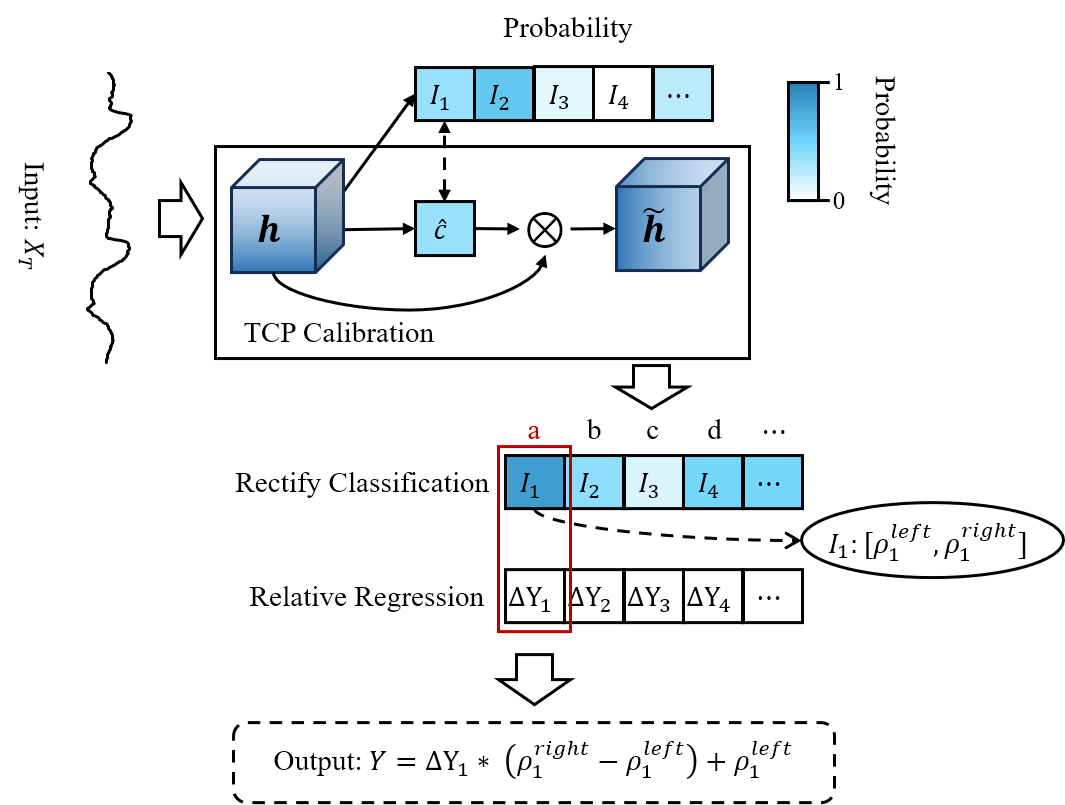}
    \caption{Probability pattern-guided time series forecasting (PPGF) strategy. We present the classifier's probability output, the TCP calibration process, and the relative prediction of the predictor.}
    \label{fig:CGR Strategy}
\end{figure}

In summary, according to Section~\ref{sec:Temporal Information}, the general feature $g$ can be obtained. In this section, we first adopt a fully-connected layer to get the feature vector: $h=FC_1(g) \in \mathbb{R}^{B\times dim}$, where $B$ means batch size and $dim$ means the dimension of hidden layer. 
Then, a classifier produces the classification probability $c$, and ConfidNet trained with $\mathcal L^{conf}$, obtains the confidence of the true class, $\hat{c}$. Accordingly, we can assess the reliability of the feature $h$. A gating strategy is then applied to reweight the feature vector $h$, allowing the module to obtain calibrated features: $\tilde{h}=h\odot \hat{c}$, where $\odot$ represents element-wise multiplication, as shown in Figure~\ref{fig:CGR Strategy}.

Finally, an additional classifier $FC_2$ is used to obtain the final classification results: $k=FC_2(\tilde{h})$ according to the calibrated features, and the classifier is trained with cross-entropy loss to minimize the Kullback-Leibler divergence between the predictive distribution and the actual distribution:
\begin{align}
\label{formula:loss}
\mathcal L^{cls} &= \frac{1}{N}\sum_{i}\mathcal L_i \nonumber \\
&=-\frac{1}{N}\sum_{i}^{N}k_i log(\omega_i h\odot \hat{c}_i +b),
\end{align}
where $\omega$ is the trainable parameter, and $b$ is the bias. We take the derivative of this loss and get:
\begin{align}
\label{formula:partial}
\frac{\partial \mathcal L^{cls}}{\partial \omega_i} 
&= -\frac{1}{N}\sum_{i}^{N} k_i \frac{1}{\omega_i h\odot \hat{c}_i + b}\times h\odot \hat{c}_i \nonumber \\
&= -\frac{1}{N}\sum_{i}^{N} k_i \frac{h}{\omega_i h + \frac{b}{\hat{c}_i}}.
\end{align}

Through the formulas \ref{formula:loss} and \ref{formula:partial}, we find that when the classifier makes a mistake, $\hat{c}$ will be small, and the wrong classified sample produces a big loss. When the classification of classifier, $c^*$, is correct, $\hat{c}$ will be large, and it produces a small loss. Therefore, the model pays more attention to these error samples for better performance.

\subsection{Relative Prediction Strategy}
\label{sec:Regression Information}
As mentioned in Section \ref{sec:problem}, we reformulate the TSF problem as the relative prediction. 

We adopt $\Delta Y$ refers to the offset of $Y$ to the lower bound of interval $k$. To predict $\Delta Y$, a network $FC_3:g\to \Delta Y$ is trained based on general features obtained by Temporal Information Extractor $g$:
\begin{align}
\Delta Y = \text{ReLU} (FC_3(g)),
\end{align}
where we denote $FC_3(g)$ as the full connection layer and adopt the ReLU activation function.

Here, the MSE loss is used to ensure the performance of the relative estimation, defined as $\mathcal L^{reg}$:

\begin{align}
\mathcal L^{reg} = (\Delta Y - {\Delta \hat{Y}})^2,
\end{align}
where $\Delta Y$ means the true label, and $\Delta \hat{Y}$ represents the predict results.

\subsection{Probability Pattern-Guided Forecasting Strategy}
\label{sec:CGR Strategy}
According to Section \ref{sec:Classification Information} and Section \ref{sec:Regression Information}, the classifier performs pattern classification after confidence correcting to determine which pattern the predicted values belong to. Then the regressor make accurate predictions in the corresponding small interval: $\Delta Y$. This section proposes the strategy of probability pattern-guided forecasting to maintain the consistency between classification and relative forecasting, as shown in Figure~\ref{fig:CGR Strategy}. 

We perform classification tasks for pattern discovery and apply forecasting tasks to the truth-value pattern interval to train the model. Specifically, when the real predicted value $y_i$ of input sequence $x_i$ belongs to the interval of the group $k$, that is, $\Delta y_k \in \mathcal{I}_k$, and the forecasting label is $\Delta y_k = \frac{y_i-\rho_k^{left}}{\rho_k^{right}-\rho_k^{left}}$.

When inference, the overall forecasting process of the proposed method can be written as follows:
\begin{align}
y_i = {\Delta y}_{k} \ast (\rho_{k}^{right}-\rho_{k}^{left})+\rho_{k}^{left},
\end{align}
where $k$ means the group with the highest probability.

\subsection{Objective Function}
\label{sec:Objective Function}
The total loss function mainly consists of three parts: confidence network loss $\mathcal L^{conf}$, coarse-grained probabilistic pattern classification loss $\mathcal L^{cls}$ and fine-grained relative prediction loss $\mathcal L^{reg}$:
\begin{align}
\mathcal L = \sum_{i=1}^N (\lambda_1 \mathcal L^{conf} + \lambda_2 \mathcal L^{cls} + \lambda_3 \mathcal L^{reg}),
\end{align}
where $\lambda_1$, $\lambda_2$ and $\lambda_3$ are hyper-parameters used to balance the influence of true label probability, classification, and forecasting. The model can be obtained by optimizing the loss $\mathcal L$. The contribution of these tasks to our current work is different. After experimental verification, this paper adopts $\lambda_1=\lambda_2=1$, $\lambda_3=5$. In a future extension of our model, we will study how to automatically learn weights for each task.

\section{Experiments}
\label{sec:evaluation}
We conducted extensive experiments with fourteen methods on eight available datasets for TSF tasks.

\subsection{Datasets and Analysis.}
\label{sec:datasets}
We used following datasets that are publicly available. Table~\ref{tab:Dataset} summarizes the dataset statistics.
\begin{enumerate}
    \item \textbf{Power Consumption (Zone1-3)}\footnote{\url{https://archive.ics.uci.edu/ml/datasets/Power+consumption+of+Tetouan+city#}}: the power consumption of Zone 1-3 distribution networks of Tetouan city located in northern Morocco.
    \item \textbf{Steel Industry Energy Consumption}\footnote{\url{https://archive.ics.uci.edu/ml/datasets/Steel+Industry+Energy+Consumption+Dataset}}: industry energy consumption in kWh collected from a smart small-scale steel industry in South Korea.
    \item \textbf{Metro Interstate Traffic Volume}\footnote{\url{https://archive.ics.uci.edu/ml/datasets/Metro+Interstate+Traffic+Volume}}: hourly Minneapolis-St Paul, MN traffic volume for westbound I-94 from 2012-2018.
    \item \textbf{Air Quality}\footnote{\url{https://archive.ics.uci.edu/ml/datasets/Air+quality}} \cite{sun2021adjusting}: recorded by gas multi-sensor devices deployed on the field in an Italian city.
    \item \textbf{Electricity Load Diagrams}\footnote{\url{https://archive.ics.uci.edu/ml/datasets/ElectricityLoadDiagrams20112014}} \cite{sun2021adjusting}: electricity consumption (kWh) of 370 clients from 2012 to 2014.
    \item \textbf{Solar Wind Speed}\footnote{\url{https://github.com/syrGitHub/GTA}} \cite{sun2022solar}: solar wind speed data set collected by NASA from 2011-2017.
\end{enumerate}

Except the solar wind data, other data sets are divided into training (60\%), validation (20\%), and testing (20\%) in chronological order. Solar wind speed data from 2011 to 2015 are used as the training set, 2016 as the verification set, and 2017 as the test set. In the data preprocessing, the data is normalized.

\begin{table}[!t]
	\caption{Dataset Statistics.}
	\label{tab:Dataset}
	\resizebox{\linewidth}{!}{
		\begin{tabular}{lrrr}
			\toprule
			Datasets             & Timesteps     & Variates   & Granularity          \\ \midrule
			Power consumption (Zone1-3)       & 52417 & 7   & 10 minutes \\
			Steel Industry Energy Consumption & 35040 & 8   & 15 minutes \\
			Metro Interstate Traffic Volume   & 40575 & 3   & 1 hour     \\
			Air Quality                       & 9358  & 15  & 1 hour     \\
			Electricity Load Diagrams         & 26304 & 321 & 15 minutes \\
			Solar Wind                        & 2545  & 1   & 1 day      \\ \bottomrule
	\end{tabular}}
\end{table}

\begin{figure}[!t]
	\centering
	\begin{minipage}[t]{0.45\linewidth}
		\centering
		\includegraphics[width=\textwidth]{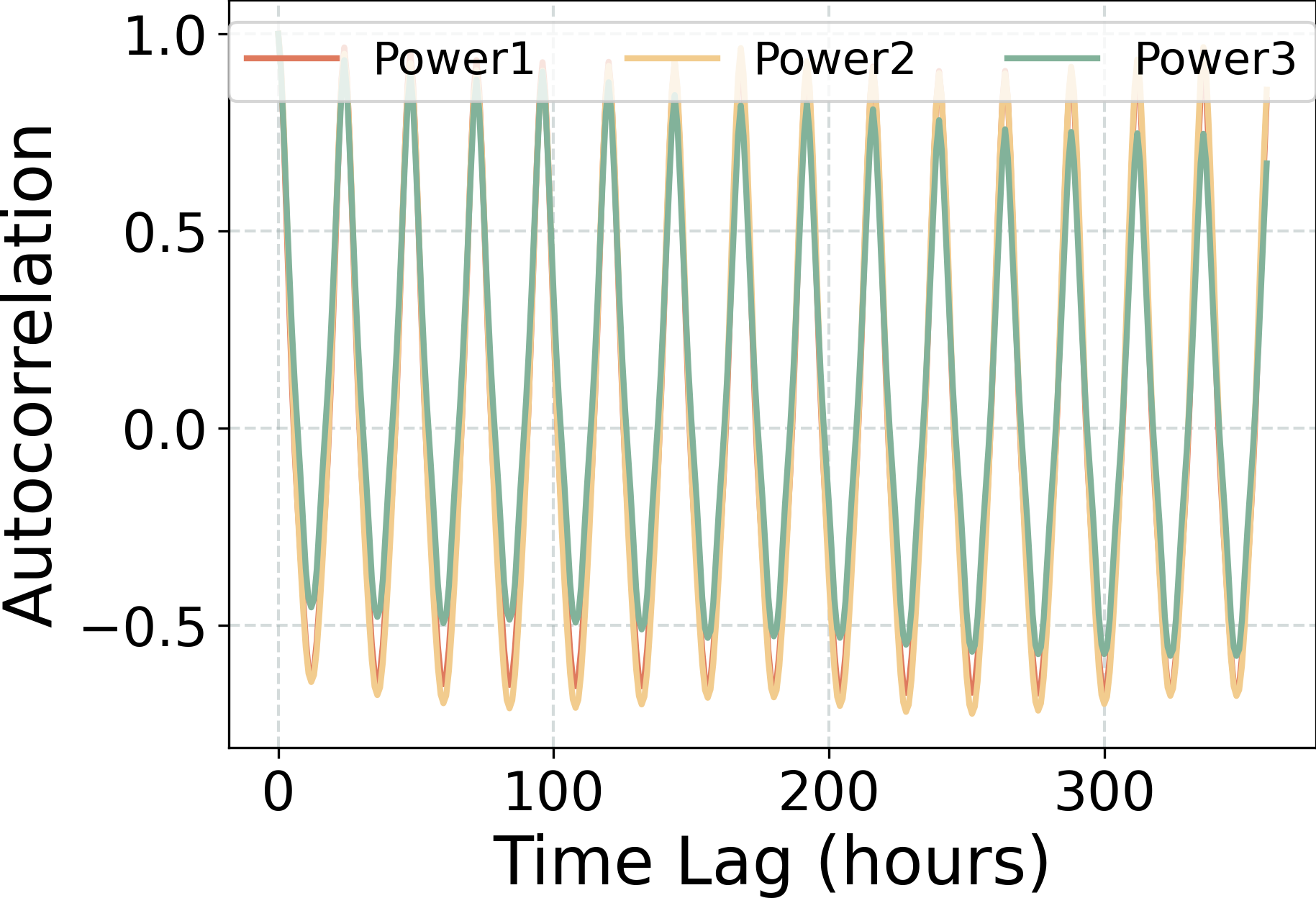}
		\centerline{(a) Power (Zone1-3)}
	\end{minipage}\hspace{0.5mm}    
	\begin{minipage}[t]{0.45\linewidth}
		\centering
		\includegraphics[width=\textwidth]{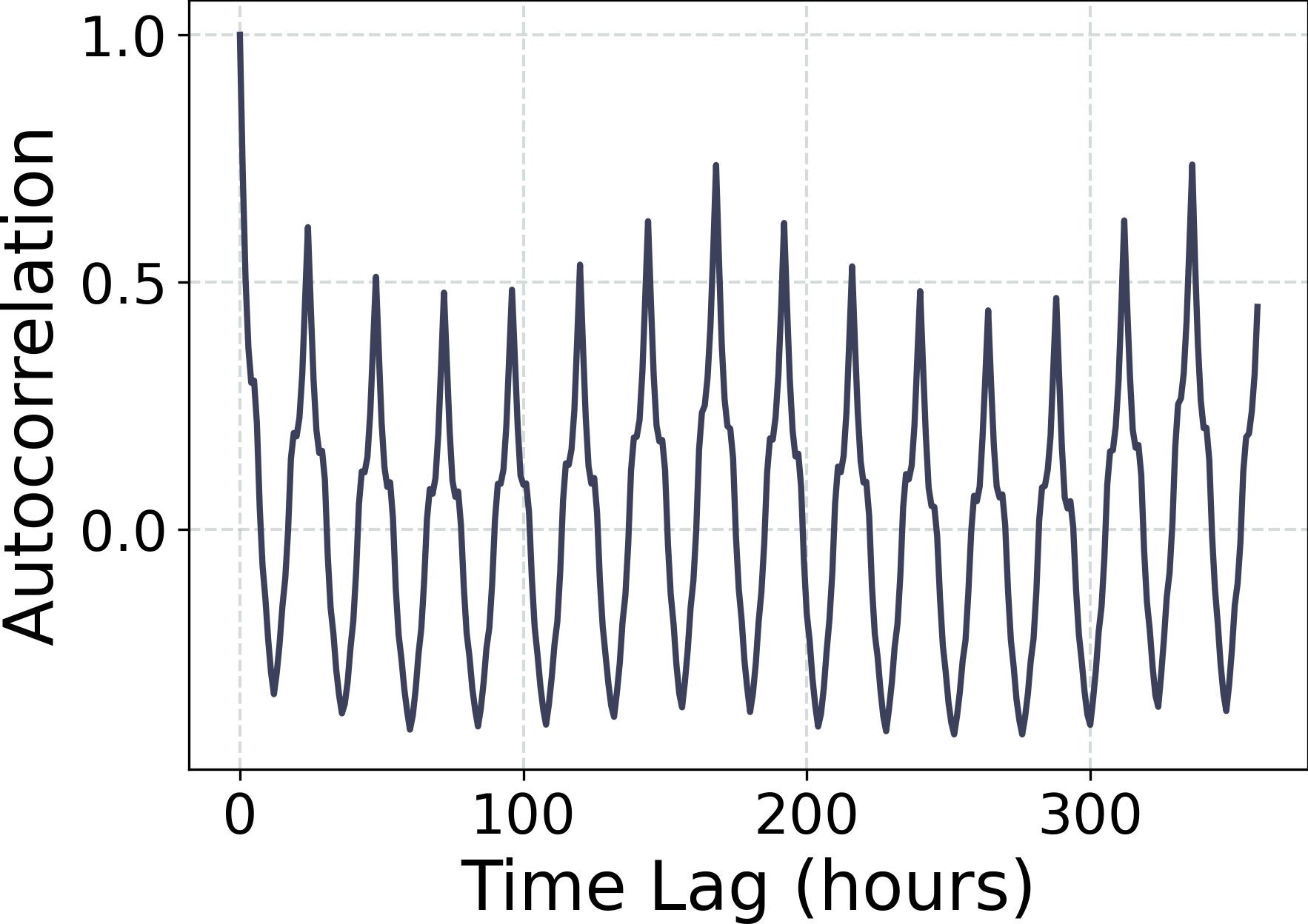}
		\centerline{(b) Steel Energy}
		\vspace{0.3mm} 
	\end{minipage}
	
	\begin{minipage}[t]{0.45\linewidth}
		\centering
		\includegraphics[width=\textwidth]{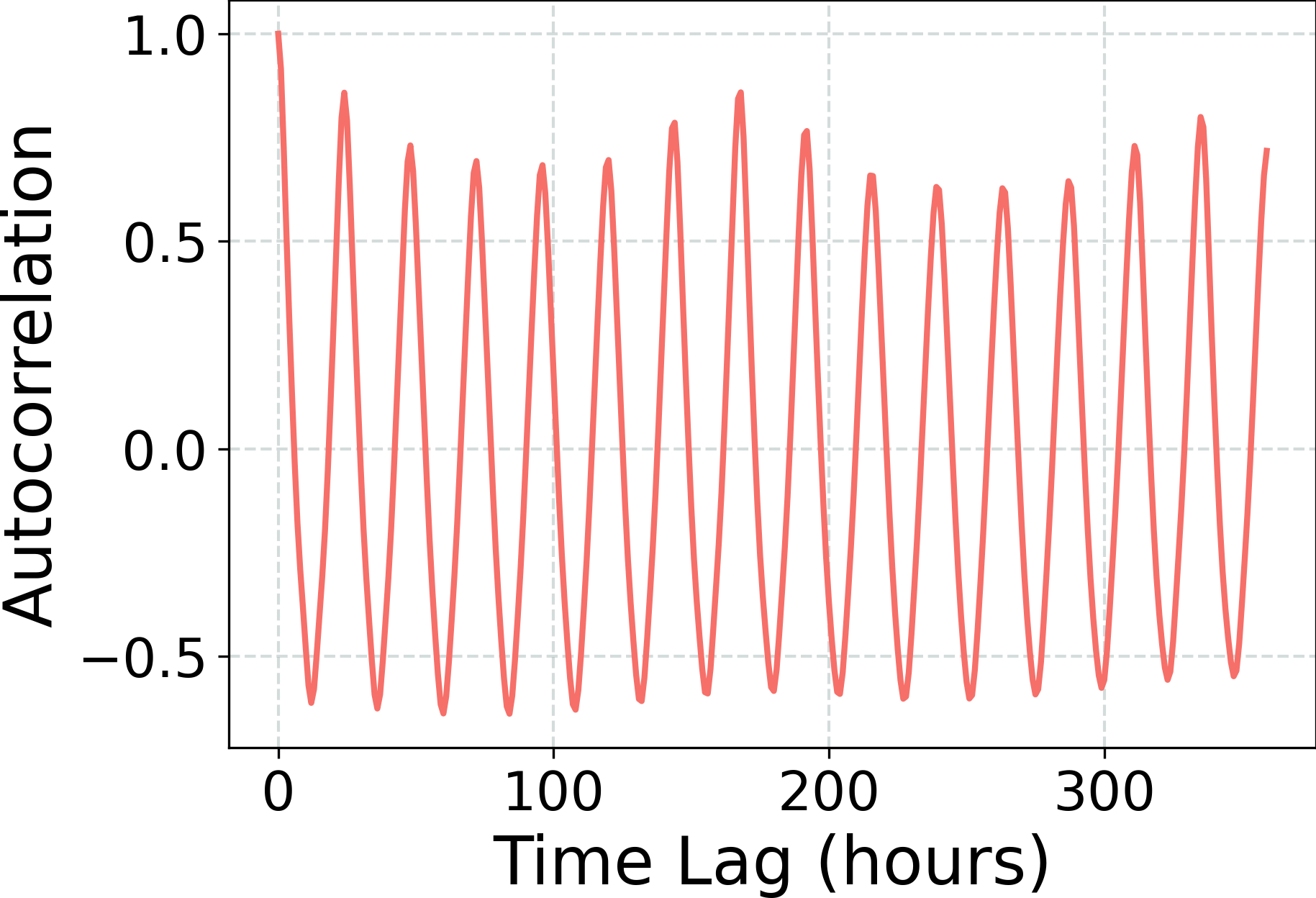}
		\centerline{(c) Traffic Volume}
	\end{minipage}\hspace{0.5mm}    
	\begin{minipage}[t]{0.45\linewidth}
		\centering
		\includegraphics[width=\textwidth]{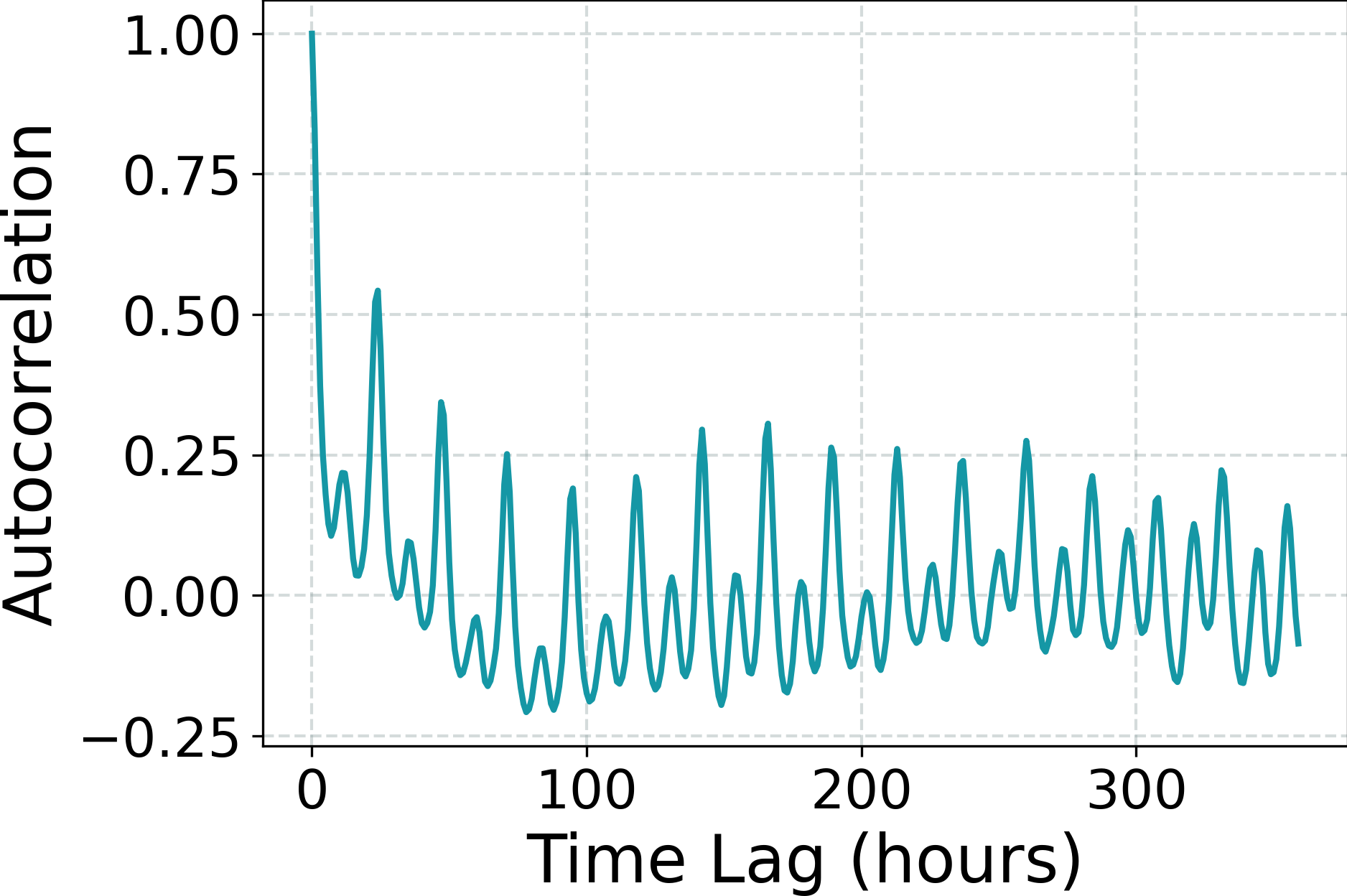}
		\centerline{(d) Air Quality}
		\vspace{0.3mm} 
	\end{minipage}
	
	\begin{minipage}[t]{0.45\linewidth}
		\centering
		\includegraphics[width=\textwidth]{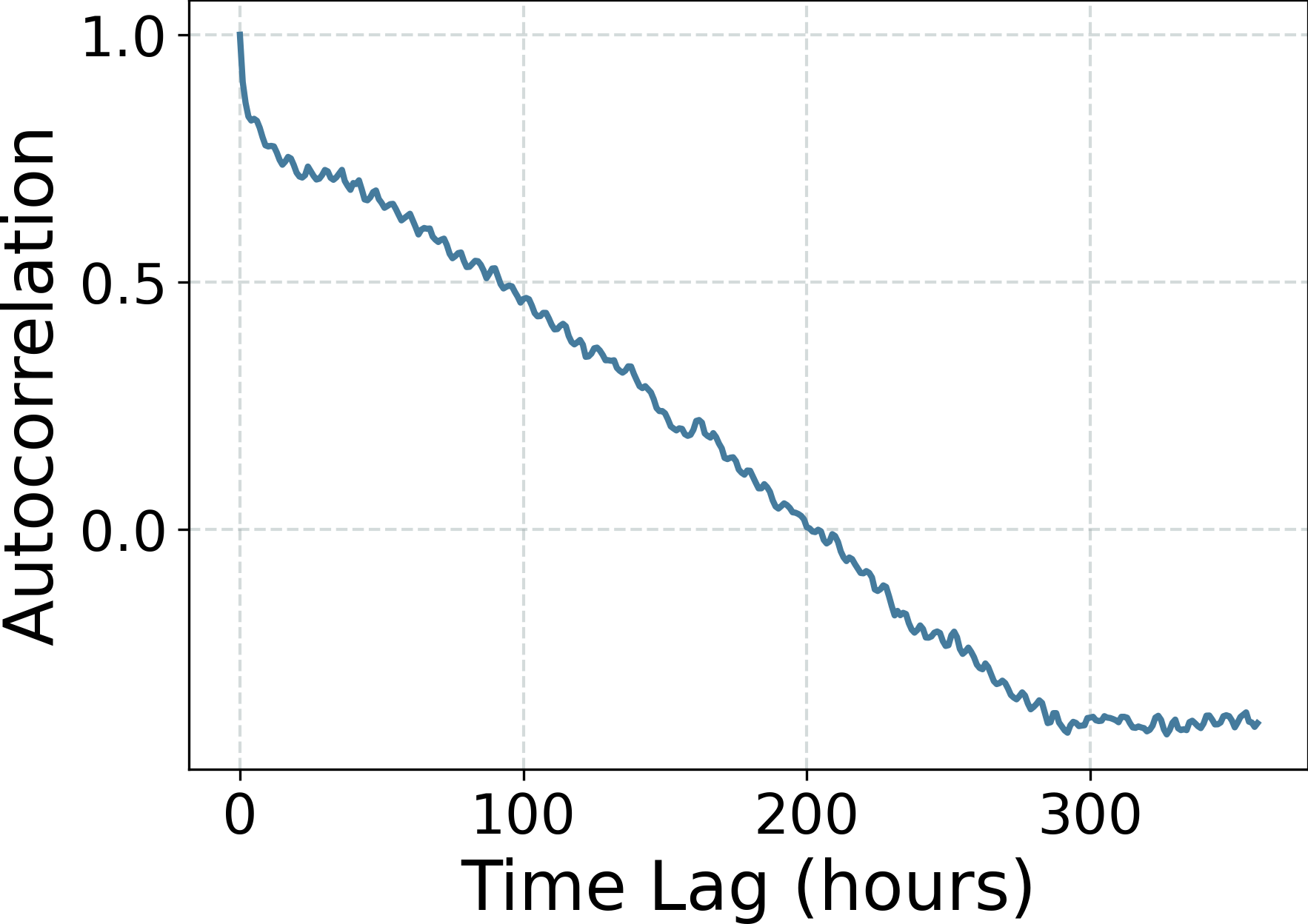}
		\centerline{(e) Electricity Load}
	\end{minipage} \hspace{0.5mm}
	\begin{minipage}[t]{0.45\linewidth}
		\centering
		\includegraphics[width=\textwidth]{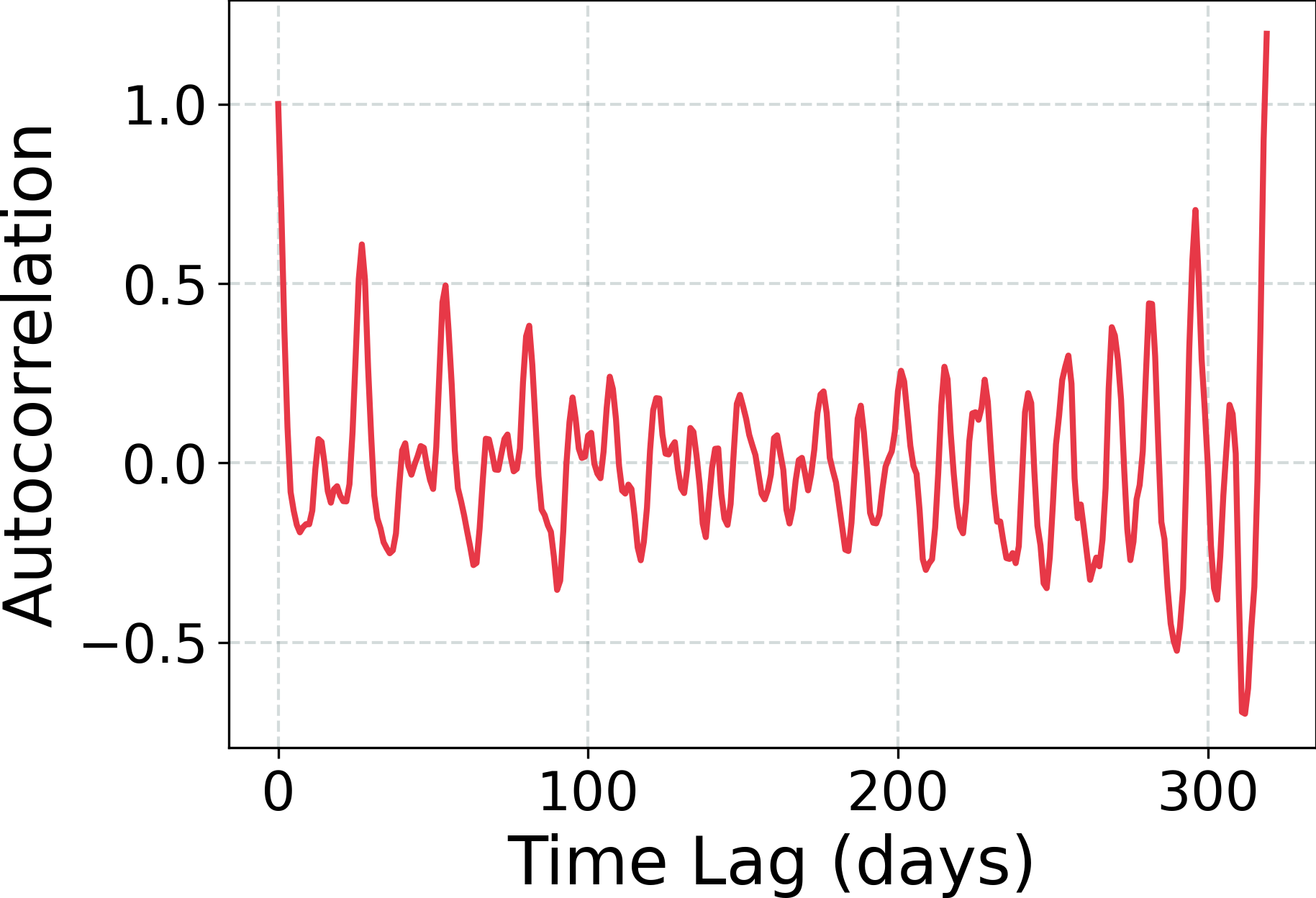}
		\centerline{(f) Solar Wind}
	\end{minipage}
	
	\caption{Autocorrelation graphs of all datasets.}
	\label{fig:Autocorrelation graphs}
\end{figure}

In order to examine the existence of repetitive patterns and periodicity in time series data, we draw autocorrelation graphs for the target variables of each dataset in Figure \ref{fig:Autocorrelation graphs}. The definition is as follows:
\begin{align}
R(\tau) = \frac{E(A_t-\mu)(A_{t + \tau}-\mu)}{\sigma^2},
\end{align}
where $A_t$ is the time series data, $\mu$ is mean and $\sigma^2$ is variance. 

As seen from graphs (a)\textasciitilde (c) in Figure~\ref{fig:Autocorrelation graphs}, there is a highly autocorrelated repeating pattern in the five data sets. However, it is weak in (d)\textasciitilde (f). In addition, we can observe daily and weekly patterns in the graphs of power consumption, steel energy consumption, and traffic volume data sets, which perfectly reflect the regularity in traffic situations and energy consumption. Our experiments have verified that PPGF achieves the best results in all of these data sets. We will revisit this point in Section \ref{sec:Main Results}.

\subsection{Experimental Setup}
\textbf{Model Comparison and Experiment Setting.}
To demonstrate the validity of our approach, we compare it against a diverse set of state-of-the-art forecasting models:
\textbf{(1) Traditional time series forecasting models:} LSTM \cite{graves2012long}, TPA \cite{shih2019temporal}, MRN \cite{guo2023multivariate}, AGCRN \cite{bai2020adaptive}, TCN \cite{bai2018empirical}, DSANet \cite{huang2019dsanet}, Conv-T \cite{li2019enhancing}, PatchTST \cite{nie2022time}, DLinear \cite{zeng2023transformers}, and Novel Seq2seq \cite{lu2024novel};
\textbf{(2) Simultaneous classification and forecasting models:} DXterMM \cite{abilasha2022deep}. The two-step method indicates that PPGF trains the classifier and the regressor separately, while the multi-task strategy allows PPGF to complete classification and forecasting tasks end-to-end.

\textbf{Experiments Details.}
We conduct a grid search over all hyper-parameters on the validation set for each dataset. Specifically, we employ grid search to train PPGF with the learning rate = $\{0.0001, 0.0005, 0.001, 0.005\}$, number of pattern = $\{2, 3, 4, 8\}$, $\lambda_1=\{1, 2, 3, 4, 5\}$, $\lambda_2=\{1, 2, 3, 4, 5\}$, and $\lambda_3=\{1, 2, 3, 4, 5\}$. Furthermore, we set the batch size to 32 and trained with Adam~\cite{KingmaB2015adam} optimizer. For Solar Wind Dataset, the length of the look-back window $L$ is 27 according to the Sun periodicity, and for other data sets, the length of the look-back window is 32. To ensure the fairness of the experiment, we reran DXtreMM, PatchTST, DLinear according to original papers and reproduced the other deep learning models according to the hyper-parameter settings provided by Sun et al \cite{sun2021adjusting}.

\textbf{Evaluation Metrics.}
We use six evaluation metrics defined as:
(1) Classification: 
For the classification task, we mainly adopt Precision, Recall and F1 score to evaluate the performance of models.
\begin{align}
Macro-Pression=\frac{\sum_{i=1}^KPre(i)}{K},
\end{align}

\begin{align}
Macro-Recall=\frac{\sum_{i=1}^KR(i)}{K},
\end{align}

\begin{align}
Macro-F1 = \frac{1}{K}\sum_{i=1}^K{F1{(i)}}=\frac{1}{K}\sum_{i=1}^K\frac{2*Pre(i)*R(i)}{Pre(i)+R(i)},
\end{align}
where $Pre$ and $R$ mean Precision and Recall, respectively. Higher is better for all three metrics.

(2) Forecasting:
For the forecasting task, we mainly use RMSE, MAE and CORR to evaluate the model's overall performance.
\begin{align}
RMSE = \sqrt{\frac{1}{N}\sum_{i=1}^N(y_i-y_i^*)^2},
\end{align}

\begin{align}
MAE = \frac{1}{N}\sum_{i=1}^N\lvert y_i-y_i^*\rvert,
\end{align}

\begin{align}
CORR = \frac{\sum_{i=1}^{N}(y_i^*-\bar{y_i^*})(y_i-\bar{y_i})}{\sqrt{\sum_{i=1}^{N}(y_i^*-\bar{y_i^*})}\sqrt{\sum_{i=1}^{N}(y_i-\bar{y_i})^2}},
\end{align}
where $y_i^*$ is the $i$-th predicted speed value, $y_i$ is the $i$-th observed speed value, $ \bar{y_i^*} $ and $ \bar{y_i} $ are the mean value of the predicted and observed speed value respectively, and $N$ is the number of samples. For RMSE and MAE, the lower value is better, but for CORR, the higher value is better.

\begin{table*}[t]
\renewcommand{\arraystretch}{1.2}
\centering
\caption{Performance comparison with benchmark models. The best results are highlighted in \textcolor{red}{\textbf{bold}}.}
\label{tab:Main result}
\scalebox{0.75}{
\begin{tabular}{ccc|cccccccccc|cccc}
\toprule
\multicolumn{2}{c|}{\multirow{2}{*}{Datasets/Models}}                                                                 & \multirow{2}{*}{Metrics} & \multicolumn{10}{c|}{Traditional Time Series Forecasting}                                                                                                                                              & \multicolumn{4}{c}{Simultaneous Classification and Forecacsting}  \\ \cline{4-17} 
\multicolumn{2}{c|}{}                                                                                                        &                          & LSTM & TPA  & MRN & AGCRN & TCN        & DSANet     & Conv-T     & PatchTST   & DLinear    & \begin{tabular}[c]{@{}c@{}}Novel\\ Seq2seq\end{tabular} & DXterMM      & Two-Step   & Multi-Task & PPGF        \\ \midrule
\multicolumn{1}{c|}{\multirow{9}{*}{\rotatebox{90}{\begin{tabular}[c]{@{}c@{}}Power\\ Consumption\end{tabular}}}} & \multicolumn{1}{c|}{\multirow{3}{*}{\rotatebox{90}{Zone 1}}} & RMSE                     & 1063.08 & 4833.56 & 855.93 & 2257.54  & 881.72        & 924.25        & 1250.79       & 845.80        & 971.75        & 813.90         & \textcolor{red}{\textbf{528.28}} & 992.96        & 4829.22           & 790.18          \\
\multicolumn{1}{c|}{}                                                                             & \multicolumn{1}{c|}{}                         & MAE                      & 864.79  & 3943.29 & 630.51 & 1757.11  & 489.56        & 682.45        & 943.78        & 640.18        & 624.09        & 543.44                                                     & \textcolor{red}{\textbf{382.07}} & 700.97        & 2640.23       & 532.56          \\
\multicolumn{1}{c|}{}                                                                             & \multicolumn{1}{c|}{}                         & CORR                     & 0.98    & 0.73    & 0.98   & 0.93     & \textcolor{red}{\textbf{0.99}} & 0.98          & 0.98          & \textcolor{red}{\textbf{0.99}} & \textcolor{red}{\textbf{0.99}} & \textcolor{red}{\textbf{0.99}}                                              & \textcolor{red}{\textbf{0.99}}   & \textcolor{red}{\textbf{0.99}} & 0.93          & \textcolor{red}{\textbf{0.99}}   \\ \cline{2-17} 
\multicolumn{1}{c|}{}                                                                             & \multicolumn{1}{c|}{\multirow{3}{*}{\rotatebox{90}{Zone 2}}} & RMSE                     & 939.21  & 895.43  & 541.05 & 2997.82  & 750.27        & 860.23        & 626.79        & 501.93        & 539.32        & 600.26                                                     & 514.64          & 831.11        & 5633.07       & \textcolor{red}{\textbf{439.77}} \\
\multicolumn{1}{c|}{}                                                                             & \multicolumn{1}{c|}{}                         & MAE                      & 731.81  & 677.85  & 436.43 & 2509.59  & 549.62        & 649.74        & 464.88        & 333.19        & 324.14        & 450.28                                                     & 318.84          & 543.03        & 2744.52       & \textcolor{red}{\textbf{317.48}} \\
\multicolumn{1}{c|}{}                                                                             & \multicolumn{1}{c|}{}                         & CORR                     & 0.98    & 0.98    & 0.98   & 0.92     & 0.98          & 0.98          & 0.99          & 0.99          & 0.99          & 0.99                                                       & \textcolor{red}{\textbf{1.00}}   & 0.99          & 0.90          & \textcolor{red}{\textbf{1.00}}   \\ \cline{2-17} 
\multicolumn{1}{c|}{}                                                                             & \multicolumn{1}{c|}{\multirow{3}{*}{\rotatebox{90}{Zone 3}}} & RMSE                     & 1545.45 & 898.96  & 834.63 & 2327.14  & 1900.30       & 1029.88       & 2315.99       & 784.21        & 819.61        & 820.40                                                     & 708.76          & 944.87        & 3563.93       & \textcolor{red}{\textbf{699.02}} \\
\multicolumn{1}{c|}{}                                                                             & \multicolumn{1}{c|}{}                         & MAE                      & 1096.43 & 728.67  & 636.36 & 1715.92  & 1428.51       & 762.55        & 1902.19       & 584.83        & 549.12        & 586.66                                                     & \textcolor{red}{\textbf{362.95}} & 498.66        & 1783.89       & 441.18          \\
\multicolumn{1}{c|}{}                                                                             & \multicolumn{1}{c|}{}                         & CORR                     & 0.94    & 0.97    & 0.94   & 0.78     & 0.89          & 0.97          & 0.82          & 0.97         & 0.97          & 0.97                                                      & \textcolor{red}{\textbf{0.98}}   & 0.96          & 0.88          & \textcolor{red}{\textbf{0.98}}   \\ \midrule
\multicolumn{2}{c|}{\multirow{3}{*}{\begin{tabular}[c]{@{}c@{}}Steel Industry\\ Energy \\ Consumption\end{tabular}}}            & RMSE                     & 11.49   & 11.73   & 11.54  & 25.15    & 12.67         & 12.16         & 25.56         & 11.44         & 11.71         & 11.94                                                      & 17.36           & 10.11        & 47.84         & \textcolor{red}{\textbf{9.81}}   \\
\multicolumn{2}{c|}{}                                                                                                        & MAE                      & 6.28    & 7.66    & 6.46   & 6.71     & 6.71          & 7.75          & 16.69         & 5.72          & 6.50          & 6.24                                                       & 8.24            & 6.59          & 27.28         & \textcolor{red}{\textbf{5.33}}   \\
\multicolumn{2}{c|}{}                                                                                                        & CORR                     & 0.93    & 0.93    & 0.93   & 0.91     & 0.91          & 0.93          & 0.59          & 0.93          & 0.93          & 0.93                                                       & 0.84            & 0.95          & 0.86          & \textcolor{red}{\textbf{0.96}}   \\ \midrule
\multicolumn{2}{c|}{\multirow{3}{*}{\begin{tabular}[c]{@{}c@{}}Metro Interstate\\ Traffic Volume\end{tabular}}}              & RMSE                     & 682.56  & 1145.31 & 485.03 & 544.50   & 858.50        & 863.93        & 542.33        & 450.75        & 450.93        & 486.94                                                     & 425.80          & 650.66        & 1067.66       & \textcolor{red}{\textbf{396.27}} \\
\multicolumn{2}{c|}{}                                                                                                        & MAE                      & 541.45  & 809.34  & 395.34 & 680.57   & 680.57        & 664.07        & 418.73        & 316.81        & 326.79        & 361.79                                                     & \textcolor{red}{\textbf{280.33}} & 501.87        & 858.69        & 282.04          \\
\multicolumn{2}{c|}{}                                                                                                        & CORR                     & 0.94    & 0.90    & 0.95   & 0.92     & 0.92          & 0.93          & \textcolor{red}{\textbf{0.98}} & 0.97          & 0.96          & 0.97                                                       & \textcolor{red}{\textbf{0.98}}   & 0.96          & 0.92          & \textcolor{red}{\textbf{0.98}}   \\ \midrule
\multicolumn{2}{c|}{\multirow{3}{*}{Air Quality}}                                                                            & RMSE                     & 0.93    & 0.90    & 0.86   & 1.32     & 0.97          & 1.21          & 1.36          & 0.77          & 0.77          & 0.86                                                       & 0.97            & 0.84          & 0.98          & \textcolor{red}{\textbf{0.71}}   \\
\multicolumn{2}{c|}{}                                                                                                        & MAE                      & 0.64    & 0.64    & 0.59   & 1.01     & 0.67          & 0.89          & 1.05          & 0.53          & 0.55          & 0.61                                                       & 0.64            & 0.50          & 0.78          & \textcolor{red}{\textbf{0.48}}   \\
\multicolumn{2}{c|}{}                                                                                                        & CORR                     & 0.74    & 0.75    & 0.77   & 0.24     & 0.76          & 0.66          & 0.61          & 0.83          & 0.82          & 0.78                                                       & 0.78            & 0.80          & 0.79          & \textcolor{red}{\textbf{0.86}}   \\ \midrule
\multicolumn{2}{c|}{\multirow{3}{*}{\begin{tabular}[c]{@{}c@{}}Electricity Load\\ Diagrams\end{tabular}}}                    & RMSE                     & 24.42   & 15.13   & 15.35  & 23.62    & 10.12         & 10.24         & 24.91         & 9.74          & 9.79          & 11.88                                                      & 13.63           & 16.88         & 12.67         & \textcolor{red}{\textbf{9.49}}   \\
\multicolumn{2}{c|}{}                                                                                                        & MAE                      & 19.20   & 9.81    & 8.23   & 20.73    & 6.13          & 6.97          & 22.35         & 4.86          & 5.30          & 7.44                                                       & 6.36            & 10.41         & 9.86          & \textcolor{red}{\textbf{4.22}}   \\
\multicolumn{2}{c|}{}                                                                                                        & CORR                     & 0.85    & 0.81    & 0.84   & 0.34     & 0.92          & \textcolor{red}{\textbf{0.93}} & 0.84          & 0.92          & 0.91          & 0.89                                                       & 0.85            & 0.84          & \textcolor{red}{\textbf {0.93}}          & 0.92            \\ \midrule
\multicolumn{2}{c|}{\multirow{3}{*}{\begin{tabular}[c]{@{}c@{}}Solar Wind\\ Speed\end{tabular}}}                             & RMSE                     & 97.51   & 86.11   & 84.23  & 89.81    & 78.79         & 87.88         & 101.77        & 80.36         & 78.80         & 98.87                                                      & 84.79           & 76.16         & 129.85        & \textcolor{red}{\textbf{69.51}}  \\
\multicolumn{2}{c|}{}                                                                                                        & MAE                      & 79.92   & 67.62   & 71.04  & 73.73    & 59.54         & 66.63         & 84.78         & 63.41         & 60.99         & 76.54                                                      & 62.40           & 64.31         & 107.22        & \textcolor{red}{\textbf{52.49}}  \\
\multicolumn{2}{c|}{}                                                                                                        & CORR                     & 0.62    & 0.55    & 0.58   & 0.49     & 0.69          & 0.65          & 0.51          & 0.56          & 0.66          & 0.69                                                       & 0.66            & 0.75          & 0.64          & \textcolor{red}{\textbf{0.77}}  \\ \hline
\rowcolor{pink!50} \multicolumn{3}{c|}{$1^\text{st}$}                                                                                                                                                                                               & 0       & 0       & 0      & 0        & 1             & 1                                                                  & 1             & 1             & 1             & 1                                                          & 8                                                                    & 1             & 0             & 19                                                                   \\ \bottomrule
\end{tabular}}
\end{table*}

\subsection{Main Results}
\label{sec:Main Results}

Table~\ref{tab:Main result} highlights the consistent superiority of PPGF across multiple datasets, securing the top performance in 19 top out of 24 experimental configurations.
In particular, PPGF reduces the RMSE by about 107\% compared to the baselines for power consumption, steel industry energy consumption, and traffic volume—datasets characterized by significant periodicity. For datasets with weak periodic patterns, such as air quality, electricity load, and solar wind, PPGF still demonstrates robust performance, achieving RMSE reductions of 39\%, 61\%, and 30\%, respectively.
These results underscore the effectiveness of our model, which we attribute to three primary design aspects. First, the Pattern-Guided Forecasting Strategy simplifies time series forecasting by introducing a grouping strategy that discretizes continuous labels into bins and forecasts values within the corresponding intervals. This approach enables PPGF to capture variations more effectively, even in the absence of pronounced periodic patterns. 
Second, the probability pattern classifier provides more reliable classification information, and when combined with the relative prediction strategy, it improves performance across diverse datasets.
Finally, PPGF's architecture is designed to leverage both local and global dependencies in the data, allowing it to adapt to different patterns and complexities inherent in time series data. This flexibility is particularly advantageous for datasets where traditional models may struggle to identify and exploit underlying trends.

What's more, by comparing with DXtreMM and Two-Step PPGF, we can conclude that end-to-end is a better strategy because it considers the classification and forecasting errors jointly. And DXtreMM achieves better performance than the traditional TSF models, proving that it is essential to consider different patterns in the data. Secondly, PPGF obtains the best performance compared with the multi-task strategy, which should be attributed to the classification-guided forecasting strategy. We alleviate the data imbalanced problem and limit the target value to a small interval through the pattern classification task.

In addition, we find that DXtreMM outperforms PPGF on some datasets, especially those with large magnitudes, such as Power Consumption datasets. The reason is that DXtreMM is specifically designed for extreme event prediction, and extreme values are more common in these datasets.

\begin{figure}[!t]
		\centering
	\begin{minipage}[t]{0.4\linewidth}
		\centering
		\includegraphics[width=\textwidth]{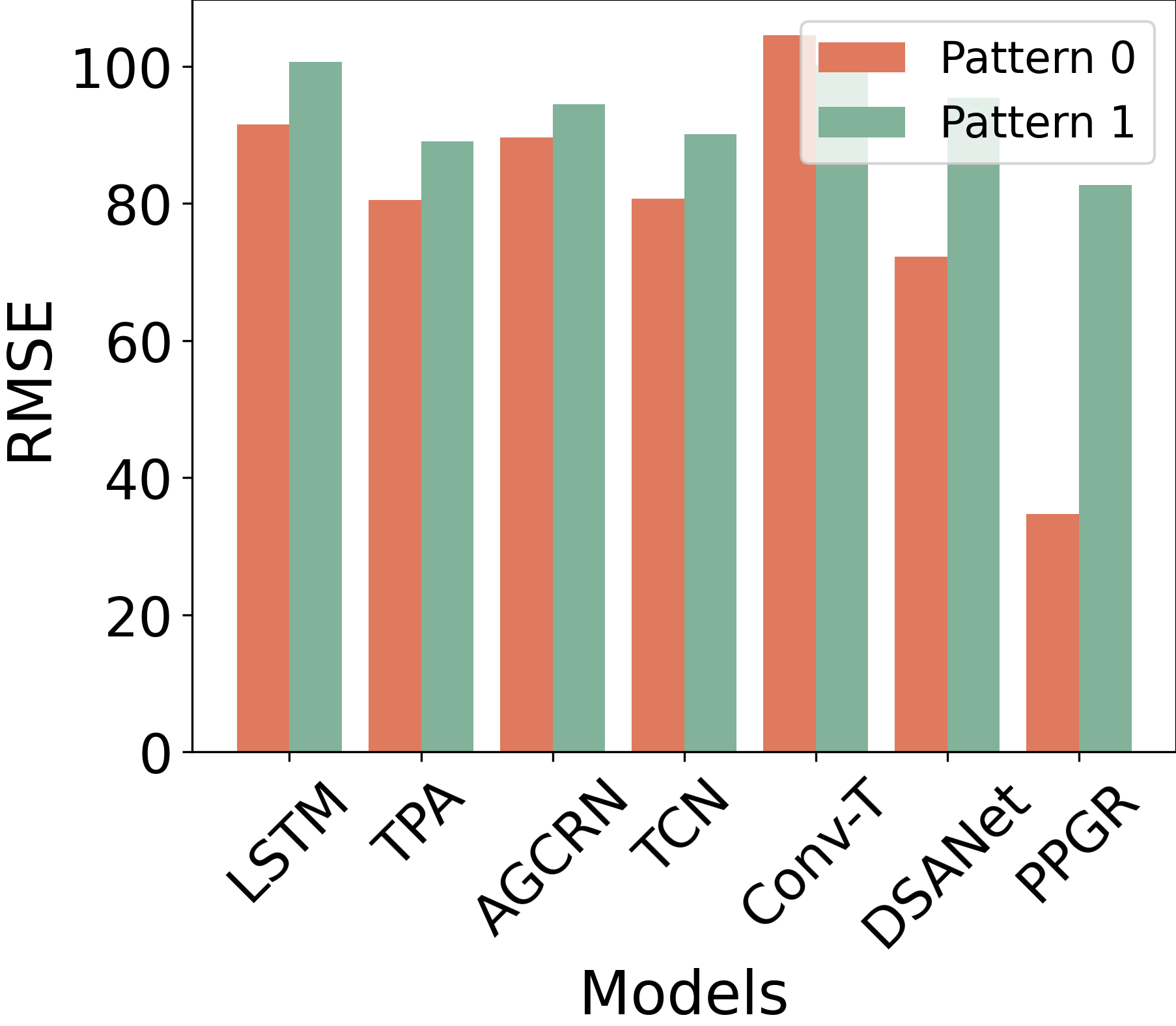}
		\centerline{(a) RMSE}
	\end{minipage}\hspace{0.5mm}
	\begin{minipage}[t]{0.4\linewidth}
		\centering
		\includegraphics[width=\textwidth]{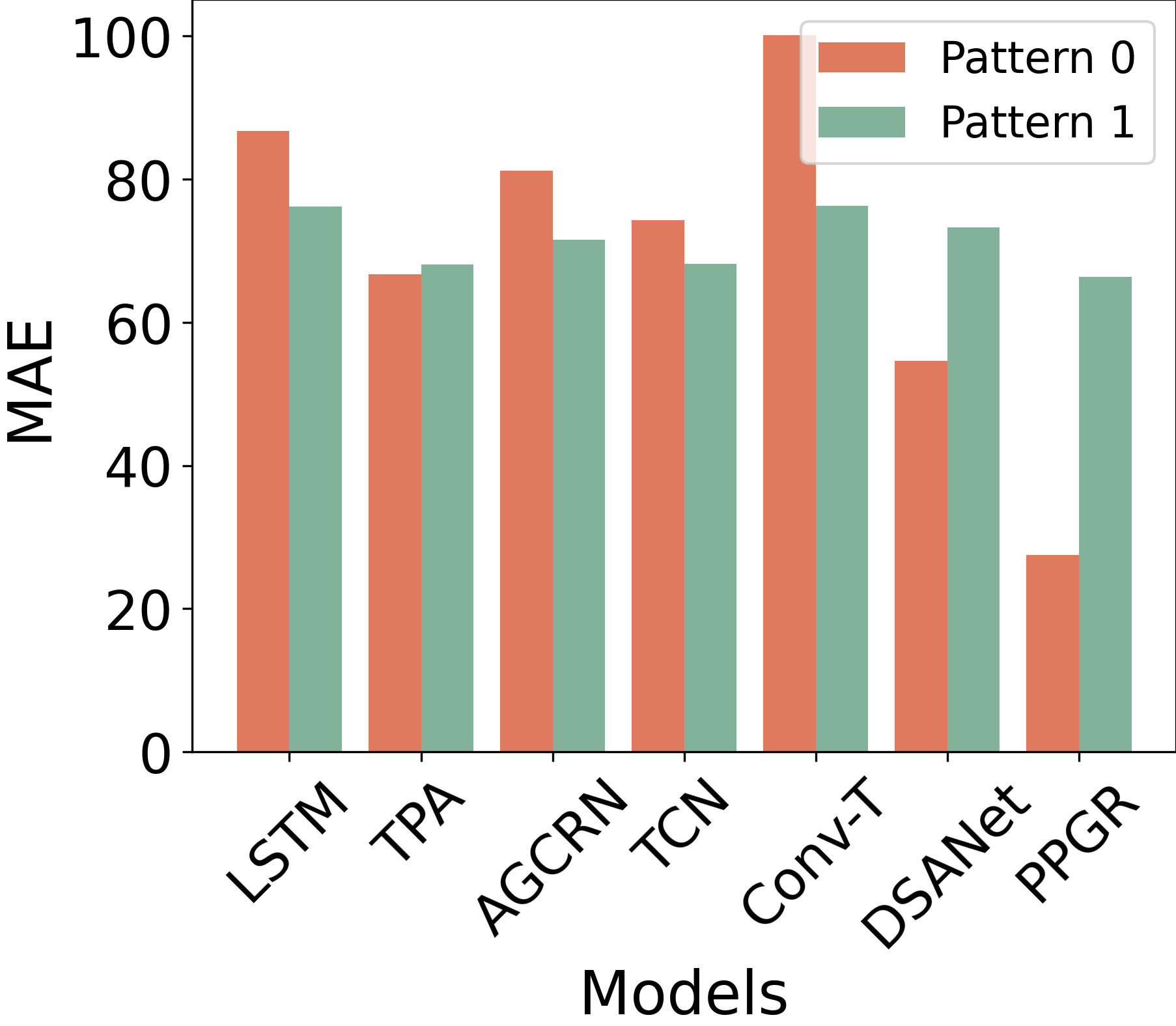}
		\centerline{(b) MAE}
		\label{fig:b}
		\vspace{-3mm} 
	\end{minipage}
	
	\begin{minipage}[t]{0.4\linewidth}
		\centering
		\includegraphics[width=\textwidth]{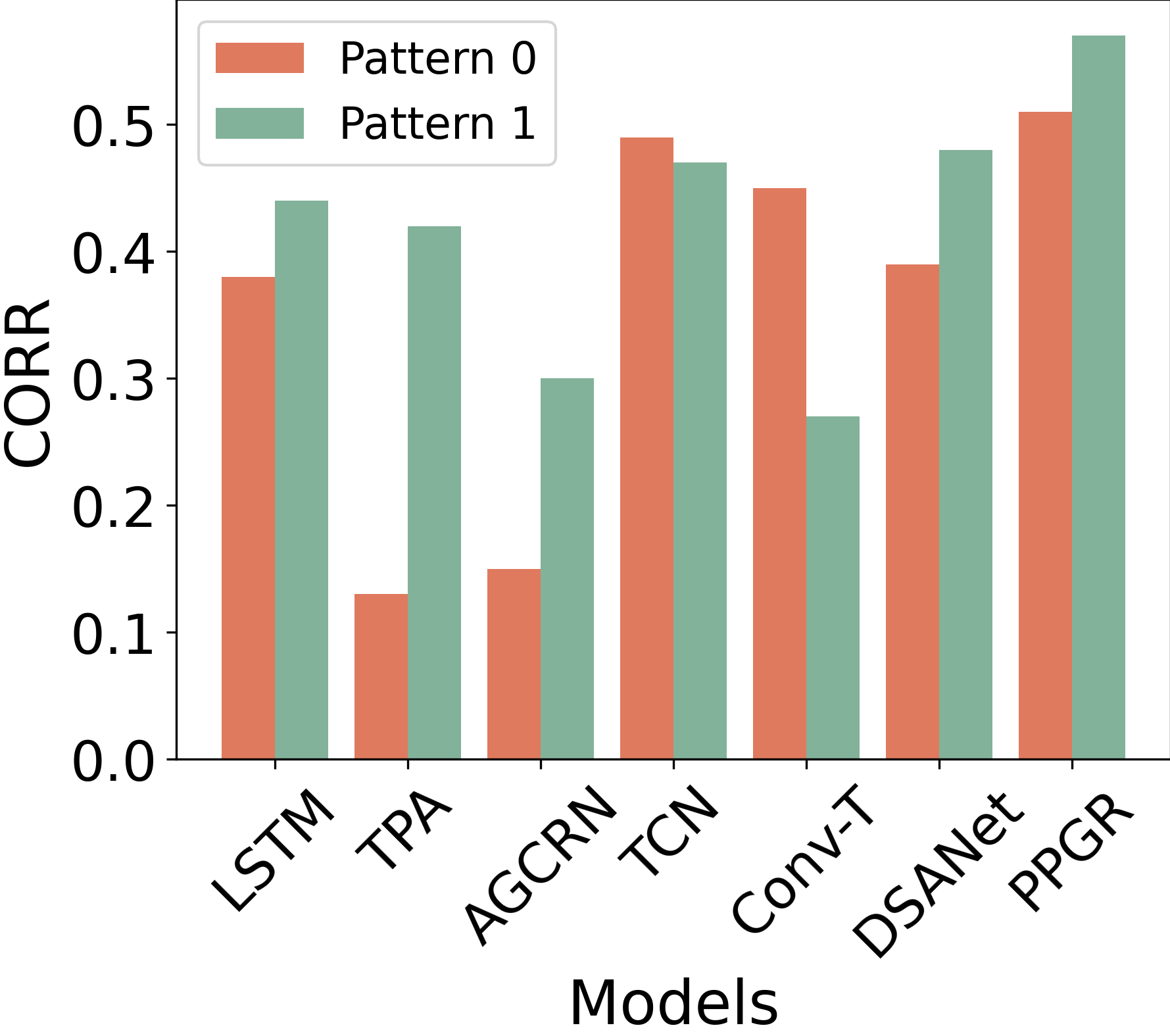}
		\centerline{(b) CORR}
	\end{minipage} \hspace{0.5mm}
	\begin{minipage}[t]{0.4\linewidth}
		\centering
		\includegraphics[width=\textwidth]{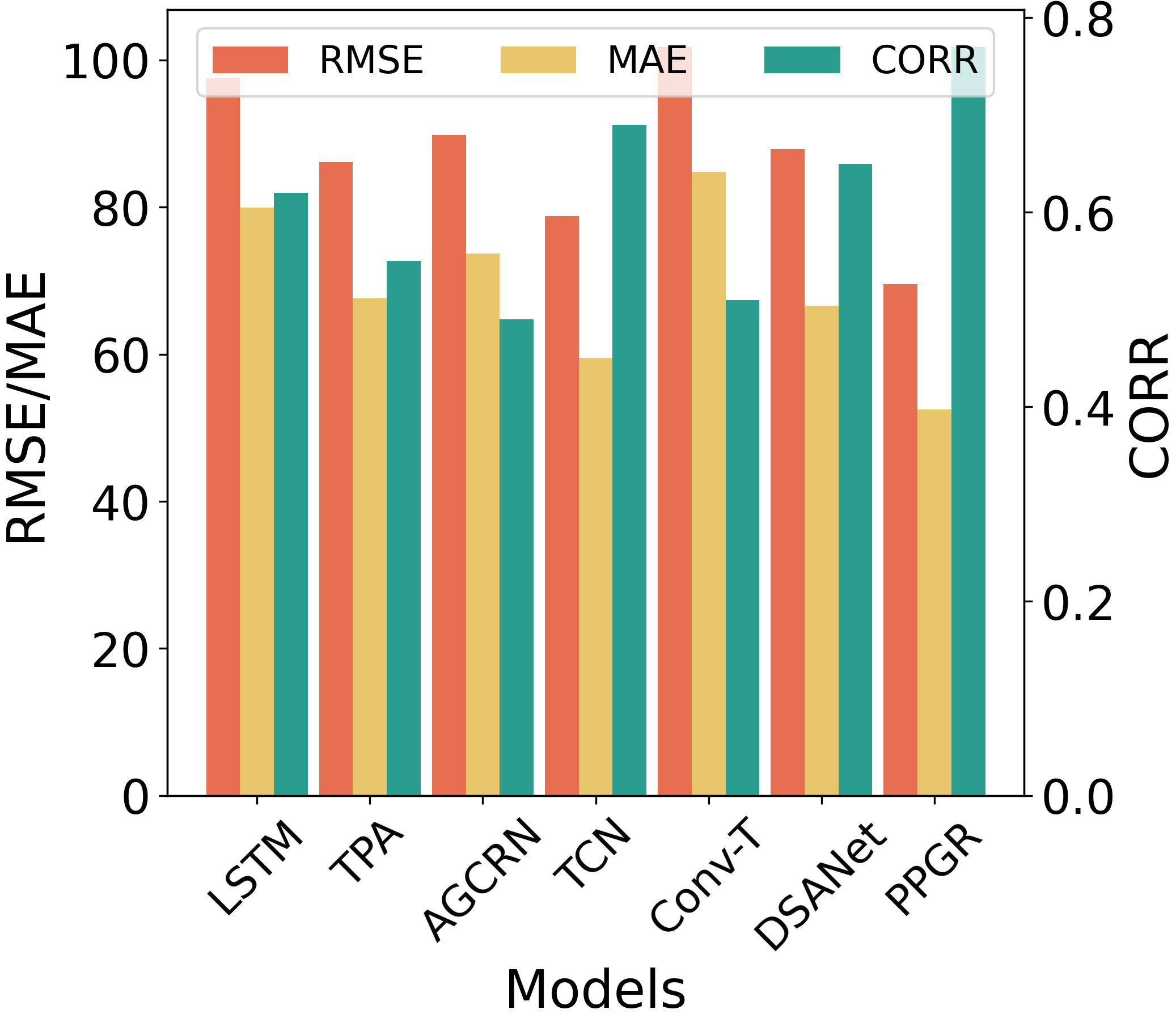}
		\centerline{(b) Overall}
	\end{minipage}
	\caption{(a) RMSE, (b) MAE, (c) CORR performance on different patterns, and (d) overall results on the solar wind dataset.}
	\label{fig:Bar Figure Result}
\end{figure}

\subsection{Classification Pattern}
This section aims to demonstrate the performance obtained for two patterns in the solar wind dataset, as well as to assess the impact of the number of patterns on the performance of various datasets. Specifically, we present the performance on the solar wind dataset under $K=2$, and investigate the effect of changing pattern numbers across different datasets.

\begin{figure}[!t]
	\centering
	\includegraphics[width=0.9\linewidth]{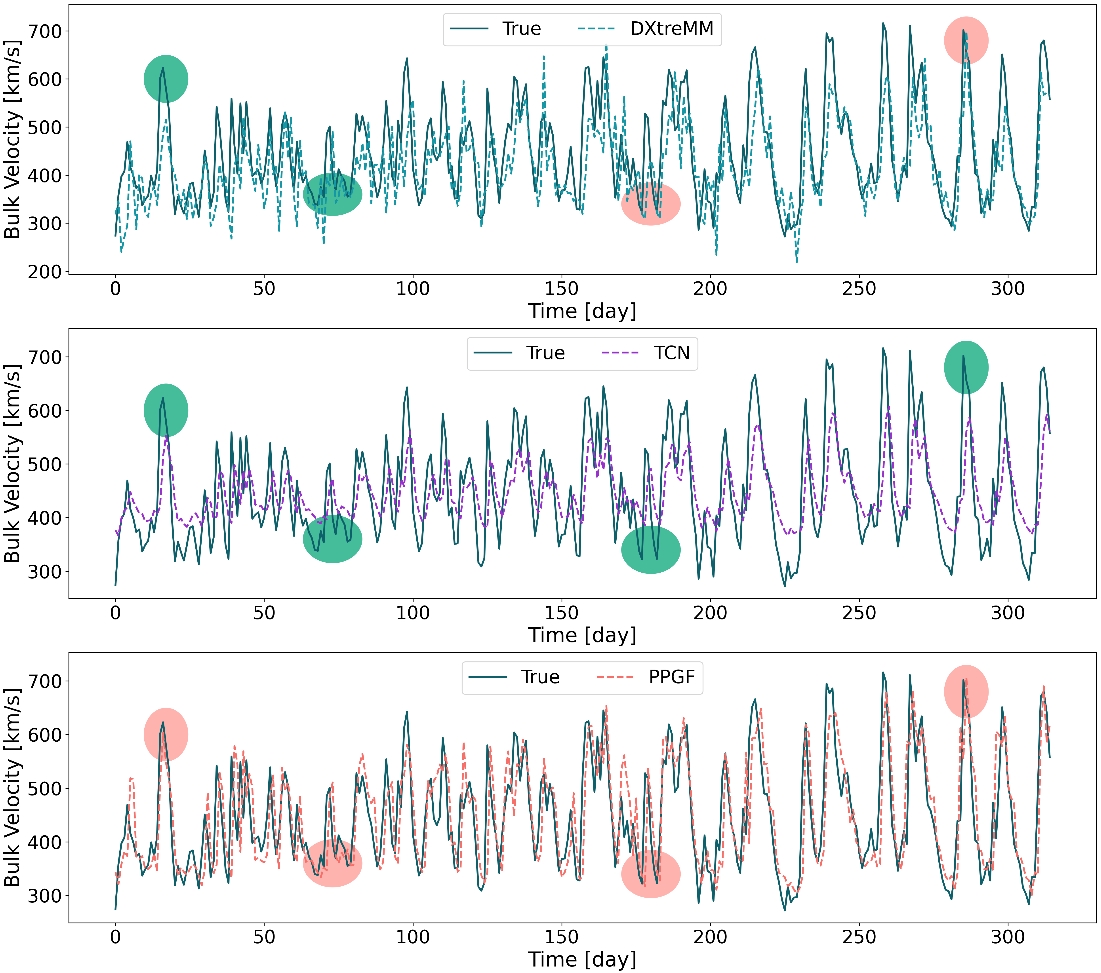}
	\caption{Visualizations of three models on the solar wind data set. Compared with other models, PPGF fits the data of different patterns well.}
	\label{fig:Visualizations}
\end{figure}

\textbf{The Results of Different Patterns.}
Coronal holes or coronal mass ejections typically cause high-speed solar wind, while low-speed solar wind generally represents the background solar wind from the Sun's surface. The two types of solar wind belong to different patterns. In Figure~\ref{fig:Bar Figure Result}, the pattern 0 means the low-speed solar wind, and the pattern 1 represents the high-speed solar wind. We can find that by introducing classification task to model different modes in different intervals, PPGF is helpful in improving the performance of each pattern. 

Futher, we visualize the prediction results of the three models on the solar wind dataset in Figure \ref{fig:Visualizations}. It can be seen that compared with other models, PPGF alleviates the problems in Figure \ref{fig:Relation} to a large extent and achieves a good fit for different patterns. We contribute these improvements to introducing the classification task, which narrows the interval of the forecasting task, reduces the difficulty of prediction, and achieves the best performance. Because DXtreMM is specially designed for extreme values prediction, it performs well at some peaks. The rest methods do not consider the category information or differentiate between different patterns, resulting in poor fit ability.

\textbf{Interpretation of the Number of Pattern $K$.}
Through experiments, we find that the number of patterns has an essential effect on the experimental results. Table~\ref{tab:The number of pattern} shows the results of different patterns for the forecasting task. 

\begin{table}[!t]
	\renewcommand{\arraystretch}{1.2}
	\centering
	\caption{RMSE of the different number of pattern in the classification information estimator module.The best results are highlighted in \textcolor{red}{\textbf{bold}}.}
	\label{tab:The number of pattern}
	\resizebox{\linewidth}{!}{
		\begin{tabular}{c|c|cccc|c}
			\toprule
			\multicolumn{2}{c|}{Number of Pattern}             & 2      & 3      & 4      & 8      & \begin{tabular}[c]{@{}c@{}}Standard\\ Deviation\end{tabular}\\ \midrule
			& Zone 1          & 889.47 & 873.42 & 877.69 & \textcolor{red}{\textbf{790.17}} & 7173.43\\
			Power Consumption      & Zone 2   & 695.11 & 787.04 & 575.38 & \textcolor{red}{\textbf{439.77}} & 4970.96\\
			& Zone 3          & 798.63 & 962.92 & \textcolor{red}{\textbf{699.01}} & 936.18 & 6464.89\\ \midrule
			\multicolumn{2}{c|}{Steel Industry Energy Consumption} & 10.2   & \textcolor{red}{\textbf{9.81}}   & 9.84   & 12.69  & 33.92\\
			\multicolumn{2}{c|}{Metro Interstate Traffic Volume}   & 458.02 & 453.15 & \textcolor{red}{\textbf{396.27}} & 494.16 & 1988.40\\
			\multicolumn{2}{c|}{Air Quality}                       & \textcolor{red}{\textbf{0.70}}   & 0.83   & 0.72   & 0.72   & 1.58\\
			\multicolumn{2}{c|}{Electricity Load Diagrams}         & \textcolor{red}{\textbf{9.49}}   & 28.23  & 26.47  & 10.01  & 24.04\\
			\multicolumn{2}{c|}{Solar Wind}                        & \textcolor{red}{\textbf{69.51}}  & 69.84  & 71.43  & 78.64  & 88.53\\ \bottomrule
	\end{tabular}}
\end{table}

We find that better results are achieved for Power Consumption Zone 1 and Zone 2 datasets when $K=8$, with RMSE of 790.17 and 439.77, respectively. For the Zone 3 and Traffic Volume datasets, the best results are achieved when $K=4$, with an RMSE of 699.01 and 396.27, respectively. For the Steel Industry Energy Consumption, better results are obtained when $K=3$, with RMSE of 9.81. For other datasets, we get the best performance with $K=2$. 

Furthermore, we analyze the relationship among the data distributions, the pattern number and the performance. As shown in Table~\ref{tab:The number of pattern}, we adopt the standard deviation of each dataset to represent the degree of imbalance. We find that when the data distribution is more imbalanced, the larger the class number $K$ is, the better effect is achieved.  
We can conclude that more detailed grouping can better alleviate the imbalance phenomenon.

\subsection{Ablation Experiments}
\label{sec:ablation}

To demonstrate the effectiveness of our framework, ablation studies are conducted. Specifically, we remove one component of PPGF at a time. First, we name the PPGF without different components as follows:
\begin{itemize}
    \item \textbf{PPGFw/oC}: the PPGF model directly forecasting without the classification.
    \item \textbf{PPGFw/oR}: the PPGF model without the Relative Prediction Strategy.
    \item \textbf{PPGFw/oConv1d}: the PPGF model without the 1D convolution component of Temporal Information Extractor.
    \item \textbf{PPGFw/oTransformer}: the PPGF model without the Transformer component of Temporal Information Extractor.
    \item \textbf{PPGFw/oGRN}: the PPGF model without the GRN component of Temporal Information Extractor.
    \item \textbf{PPGFw/oGroup}: the PPGF model adopting the "equal width" strategy to group.
    \item \textbf{PPGFw/oConf}: the PPGF model directly classify without the calibration of probability pattern classifier.
\end{itemize}

Test results measured with three forecasting metrics and three classification metrics are shown in Table~\ref{tab:Ablation Results}. A few observations from these results are worth highlighting:
\begin{itemize}
    \item The best result on each metric is obtained with PPGF.
    \item Removing either the classifier component (PPGFw/oC) or the predictor component (PPGFw/oR) from the full model leads to a significant decrease in performance, demonstrating both components' critical role and proving that the interrelationship between the two tasks plays an essential role in performance improvement.
    \item After removing Conv1d (PPGFw/oConv1d), Transformer (PPGFw/oTransformer), or GRN (PPGFw/oGRN), RMSE increases. This indicates that these components help PPGF better capture dependencies between multivariate input features and temporal dynamics, demonstrating their rationality and effectiveness.
    \item When we adopt the equal width strategy to group the samples, the classification precision obviously decreases, which results in the poor predicted performance.
    \item The removal of the ConfidNet module (PPGFw/oConf) leads to a degradation in the performance of the classification task, which in turn affects the prediction performance. The results show that the probability pattern classifier adjusts the classification features according to reliability. On the other hand, it also demonstrates that the patterns do play a guiding role in the forecasting.
\end{itemize}

The conclusion is that our architectural design is the most robust among all experimental setups.

\begin{table}[!t]
	\renewcommand{\arraystretch}{1.2}
	\centering
	\caption{Ablation results of RMSE, MAE, and CORR on Solar Wind Dataset.The best results are highlighted in \textcolor{red}{\textbf{bold}}.}
	\label{tab:Ablation Results}
	\resizebox{\linewidth}{!}{
	\begin{tabular}{ccccccc}
		\toprule
		                                                                 & \multicolumn{3}{c}{Forecasting}                                    & \multicolumn{3}{c}{MACRO-Classification}                               \\
		& RMSE                 & MAE                  & CORR                & Precision              & Recall                 & F1                   \\ \midrule
		PPGFw/oR                                                     & $--$                   & $--$                   & $--$                  & 77.82\%                & 77.62\%                & 77.72                \\
		PPGFw/oC                                                    & 78.12                & 62.07                & 0.71                & $--$                     & $--$                     & $--$                   \\
		PPGFw/oConv1d                                                    & 81.55               & 60.25               & 0.67               & 76.37\%                & 76.09\%                & 76.23                \\
        PPGFw/oTransformer                                                    & 84.74	& 64.59	& 0.61	& 70.31\%	& 69.81\%	& 70.06                \\
        PPGFw/oGRN                                                    & 85.34	& 67.65	 & 0.64	& 74.23\%	& 73.74\%	& 73.98                \\
        
        PPGFw/oGroup                                                    & 74.72                & 54.95               & 0.71                & 80.32\%                & 78.65\%                & 79.48                \\
		PPGFw/oConf                                                       & 96.92                & 70.81                & 0.58                & 76.83\%                & 76.24\%                & 76.53                \\
		PPGF                  & \textcolor{red}{\textbf{69.51}} & \textcolor{red}{\textbf{52.49}} & \textcolor{red}{\textbf{0.77}} & \textcolor{red}{\textbf{82.18\%}} & \textcolor{red}{\textbf{82.17\%}} & \textcolor{red}{\textbf{82.17}} \\ \bottomrule
	\end{tabular}}
\end{table}

\subsection{Case Study}
\begin{figure*}[!t]
    \centering
    \includegraphics[width=0.8\textwidth]{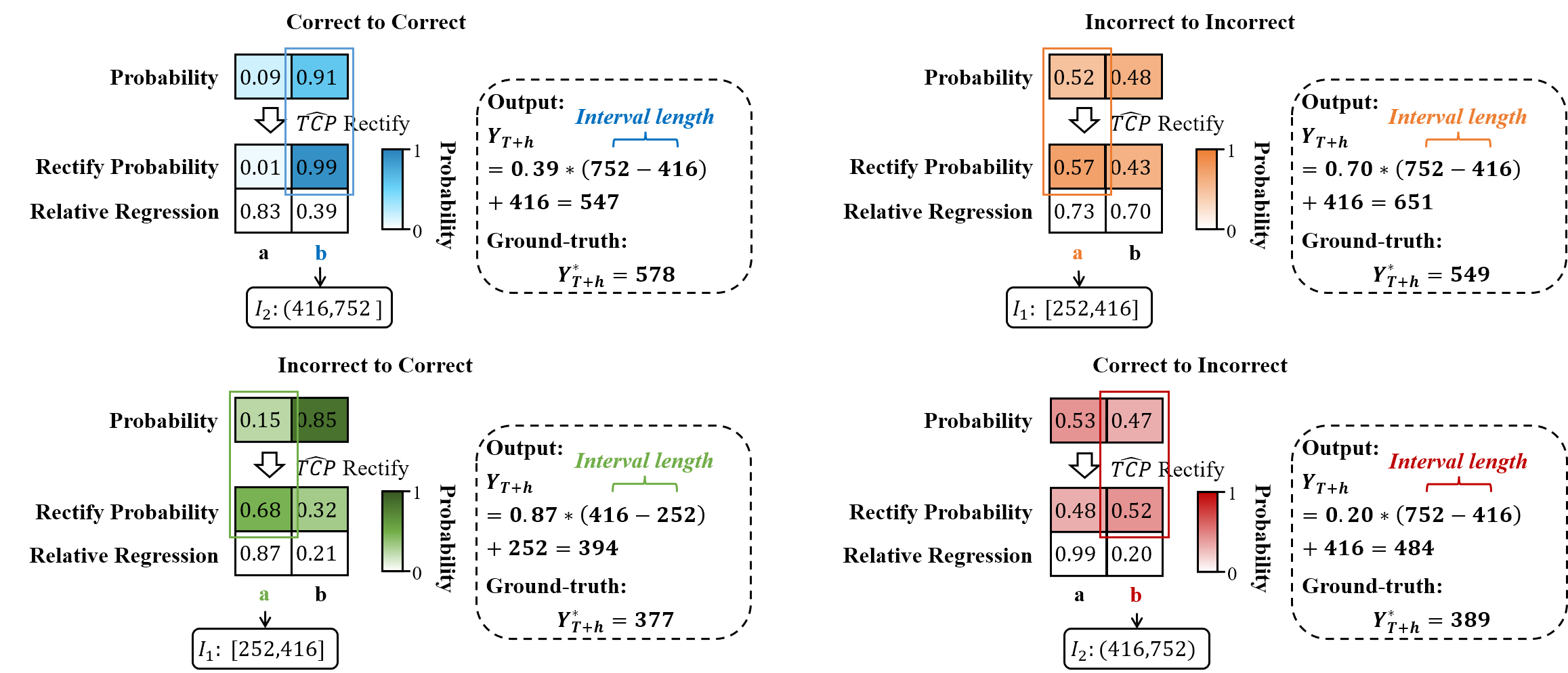}
    \caption{Case Study. We show the probability output of the pattern classifier $T$, the final classifier, and the relative prediction of the PPGF strategy. We take the prediction of the pattern with the highest probability as the final forecasting result. When calibrated successfully, the error between the predicted results and the targets is minor, proving the calibration mechanism's importance.}
    \label{fig:case}
\end{figure*}

\textbf{Model Calibration Process.}
To better understand the model's behavior, we present a case study in Figure~\ref{fig:case}. The probability pattern-guided forecasting strategy performs coarse-to-fine hierarchical forecasting based on the input samples. The probability pattern classifier first determines which pattern the target value belongs to, and the relative prediction strategy attempts to make the prediction more accurate. The four cases in the Figure~\ref{fig:case} show the result on Solar Wind Dataset when the number of pattern $K=2$. The first case shows the situation of correct classification, and the second demonstrates that when the pattern classification is wrong, it is not correct after the modification. The third case represents the samples that are misclassified but successfully calibrated. Moreover, the fourth shows the example classified correctly but falsely calibrated. From the four situations, we find that when the pattern is misclassified, the prediction will be in the wrong interval, resulting in a large error. In contrast, when the pattern is correct, it will produce a small error, proving the importance of calibration mechanism.

\begin{figure}[!t]
    \centering
    \includegraphics[width=0.8\linewidth]{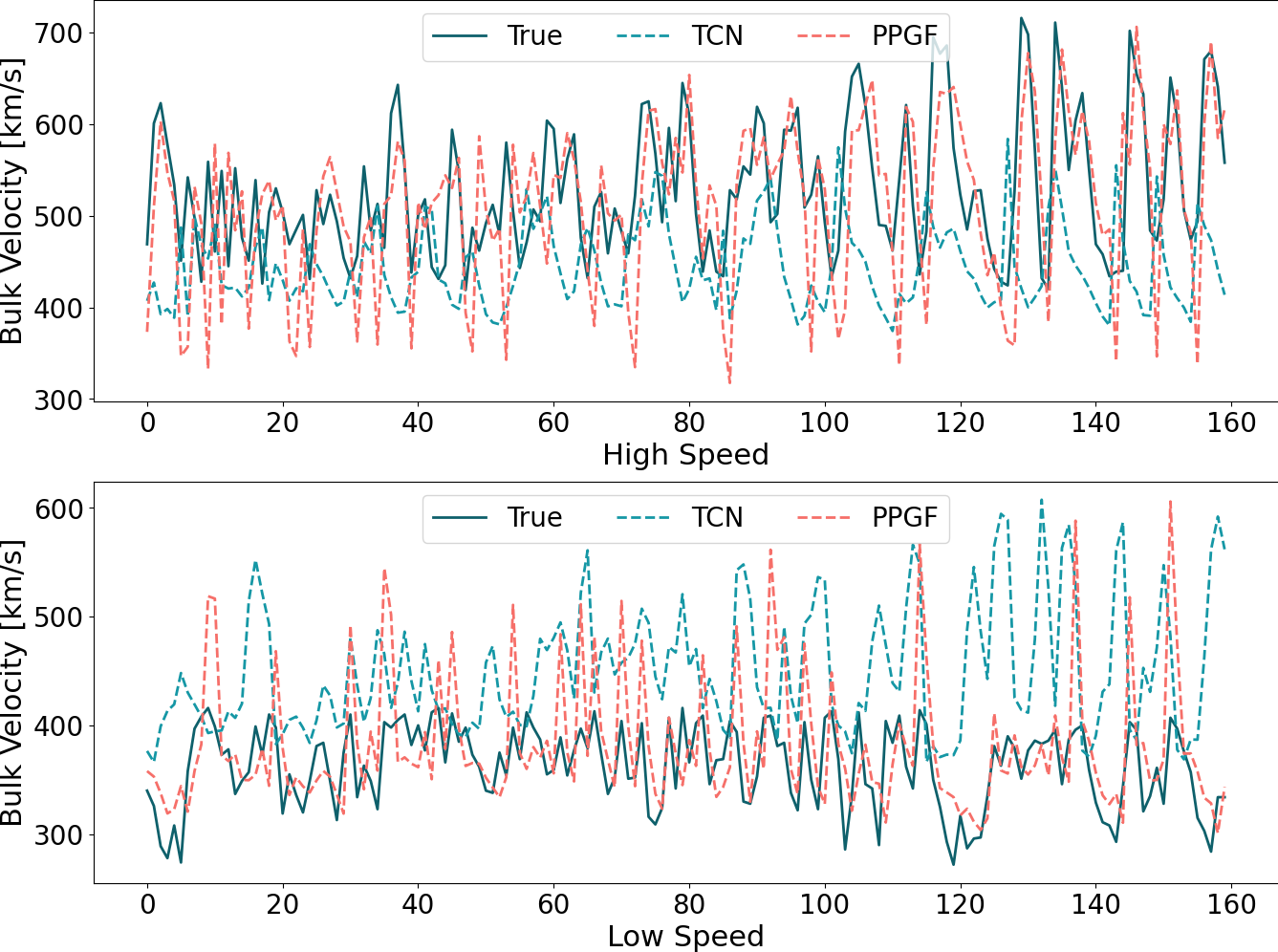}
    \caption{The predicted results of PPGF and TCN according to different patterns on Solar Wind Dataset. TCN predicts $Y$ directly. PPGF predicts the $\Delta Y$ in a compressed interval to improve the performance.}
    \label{fig:PPGR&TCN}
\end{figure}
\textbf{The Role of Pattern-Guided Forecasting.}
We draw the predicted results of PPGF and TCN according to different patterns, shown in Figure~\ref{fig:PPGR&TCN}. We find that PPGF achieves accurate performance in the parts of samples that are easy to classify correctly. As stated in Section~\ref{sec:CGR Strategy}, the predicted $Y$ with PPGF depends on the category classified by the probability pattern classifier and the offset $\Delta Y$. We conducted the MSE loss on $\Delta Y$, not on the predicted $Y$. So PPGF predicts $\Delta Y$ in a compressed range compared with TCN predicts $Y$ directly, which is helpful for forecasting.

\begin{figure}[!t]
    \centering
    \includegraphics[width=0.99\linewidth]{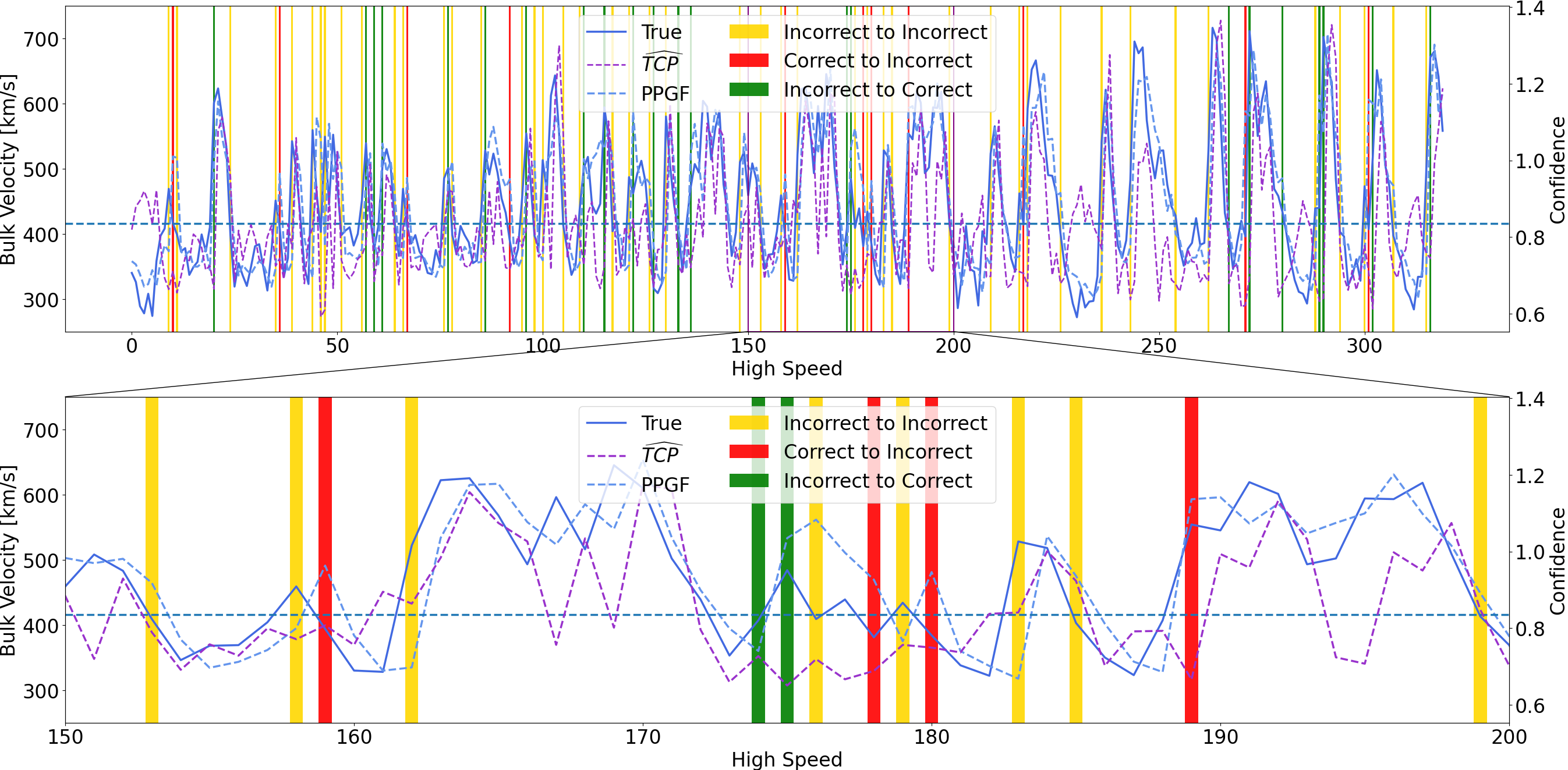}
    \caption{Predicted values, true values, and $\widehat{c}$ on Solar Wind Dataset. The yellow bars indicate that the samples are misclassified and are not calibrated. The red bars indicate that the samples are classified correctly and falsely calibrated. The green bars indicate that the samples are misclassified and successfully calibrated. The rest represents samples originally classified correctly and still classified correctly after correction.}
    \label{fig:TCP}
\end{figure}

In addition, we plot $\widehat{c}$ in Figure~\ref{fig:TCP} to visualize the function of probability pattern classifier. The yellow bars indicate that the samples are misclassified and are not calibrated. The red bars indicate that the samples are classified correctly but falsely calibrated. The green bars indicate that the samples are misclassified but successfully calibrated. The rest represents samples originally classified correctly and still classified correctly after correction. From Figure~\ref{fig:TCP}, we find that our proposed calibration mechanism is effective and successfully calibrate some misclassified samples correctly. Although some samples are calibrated incorrectly, overall, the number of samples correctly calibrated by this method is higher, which is consistent with Table~\ref{tab:Ablation Results}. In addition, we find that when the mistake occurs, $\widehat{c}$ is low. As mentioned in Section ~\ref{sec:Classification Information}, when the classifier makes a mistake, $\widehat{c}$ is small, which leads the model to pay more attention to these misclassification samples to improve performance.

\section{Conclusion}
\label{sec:conclusion}
In this paper, we propose a Probability Pattern-Guided time series Forecasting (PPGF) framework to achieve more accurate prediction. Considering the mixture of multiple patterns implicit in the data, we design a flexible pattern grouping strategy, which transforms the traditional prediction problem into pattern classification and interval relative forecasting tasks with constrained. Moreover, we introduce TCP approximation to realize the reliable pattern recognition. Experiments on real-world datasets demonstrate that the proposed approach significantly improves the state-of-the-art results in TSF on multiple benchmark datasets. Through in-depth analysis and empirical evidence, we demonstrate the PPGF model architecture's efficiency, and that it indeed successfully achieves robust prediction through reliable pattern classification.

Although the proposed method can solve the imbalance in forecasting to a degree, the pattern number is taken as a hyper-parameter which is chosen according to the results of experiments. The pattern number should be determined based on the imbalance degree of the dataset, which is our future work to realize the data-driven way in the proposed method. What's more, PPGF is prone to misclassifying samples near the boundary, affecting performance. We will pay attention to how to improve the prediction accuracy of these samples in the future. In addition, the temporal information extractor can use other advanced networks to enhance its effectiveness and applicability.


\bibliographystyle{IEEEtran}
\bibliography{reference}




\begin{IEEEbiography}[{\includegraphics[width=1in,height=1.25in,clip,keepaspectratio]{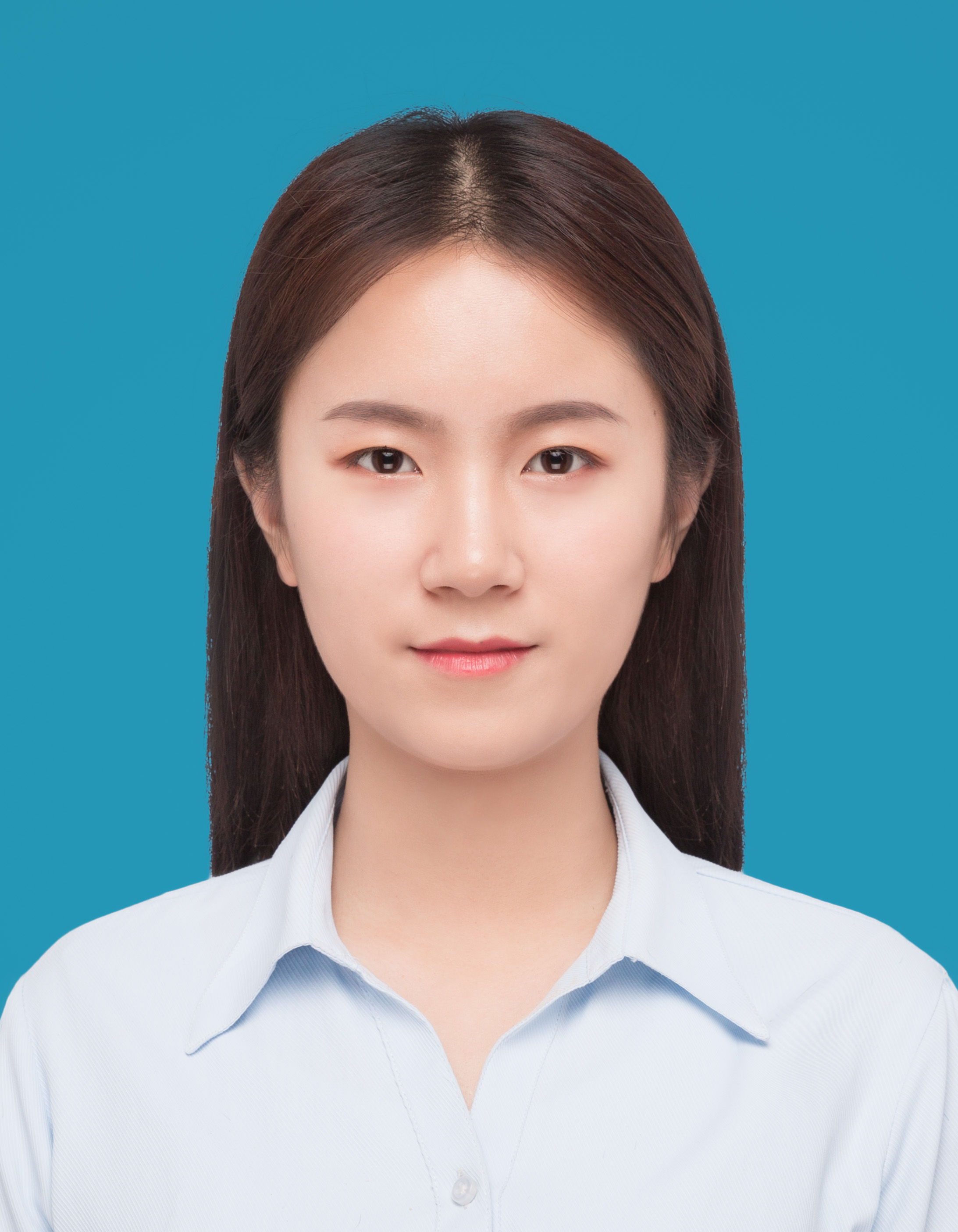}}]{Yanru Sun}
reveived the B.S. degree in computer science and technology from Shanxi University, Shanxi, China, 2015 and the M.S. degree in computer technology major from Tianjin University, Tianjin, China, 2019, respectively. She is currently working toward the Ph.D. degree in computer science and technology at the College of Intelligence and Computing, Tianjin University, Tianjin, China. Her research interests include deep learning, time series forecasting and graph neural network.
\end{IEEEbiography}

\begin{IEEEbiography}[{\includegraphics[width=1in,height=1.25in,clip,keepaspectratio]{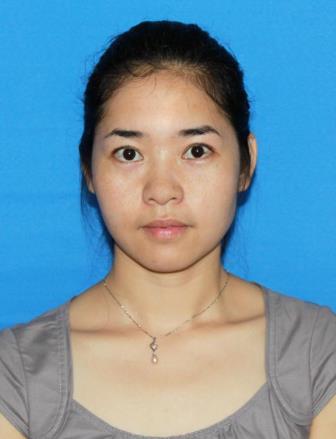}}]{Zongxia Xie}  received her B.S. from Dalian Maritime University in 2003, and M.S. and Ph.D. from Harbin Institute of Technology in 2005, and 2010, respectively. She has worked as postdoctoral researcher at Shenzhen Graduate School, Harbin Institute of Technology, from Dec. 2010 to Jan. 2013. Now she is an associate professor at College of Intelligence and Computing in Tianjin University. Her major interests include machine learning and pattern recognition.
\end{IEEEbiography}

\begin{IEEEbiography}[{\includegraphics[width=1in,height=1.25in,clip,keepaspectratio]{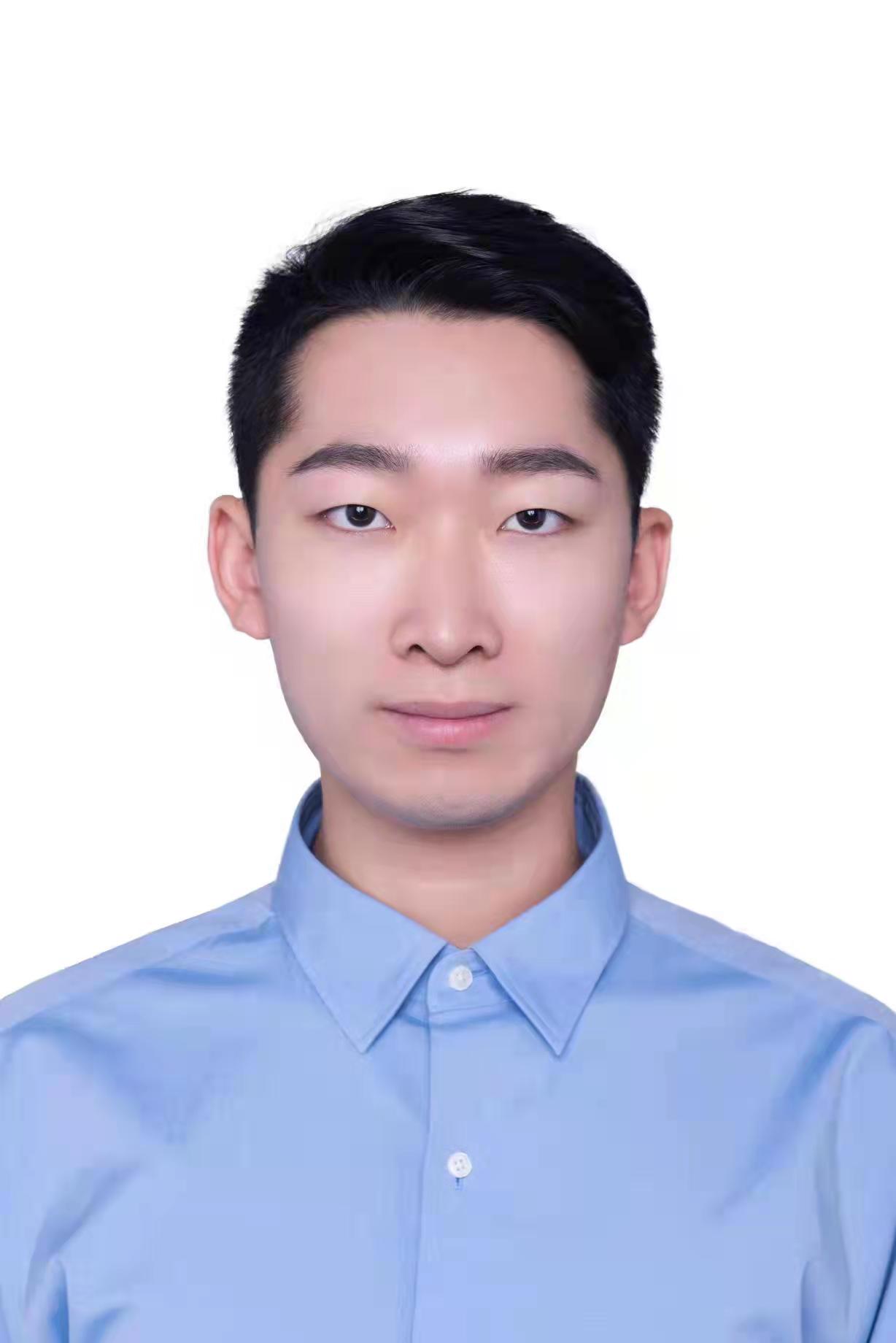}}]{Haoyu Xing}
received the B.S. degree in Computer Science from College of Intelligence and Computing, Tianjin University, Tianjin, China, in 2022. He is pursuing his M.S. degree in College of Intelligence and Computing, Tianjin University. His research interests include deep learning, time series forecasting and representation learning.
\end{IEEEbiography}

\begin{IEEEbiography}[{\includegraphics[width=1in,height=1.25in,clip,keepaspectratio]{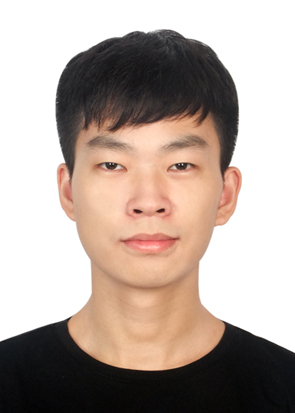}}]{Hualong Yu}
reveived the B.S. degree in software engineering from Zhengzhou University, Henan, China, 2018. He is currently working toward the M.S. degree in College of Intelligence and Computing, Tianjin University, Tianjin, China. His research interests include deep learning and time series forecasting.
\end{IEEEbiography}

\begin{IEEEbiography}[{\includegraphics[width=1in,height=1.25in,clip,keepaspectratio]{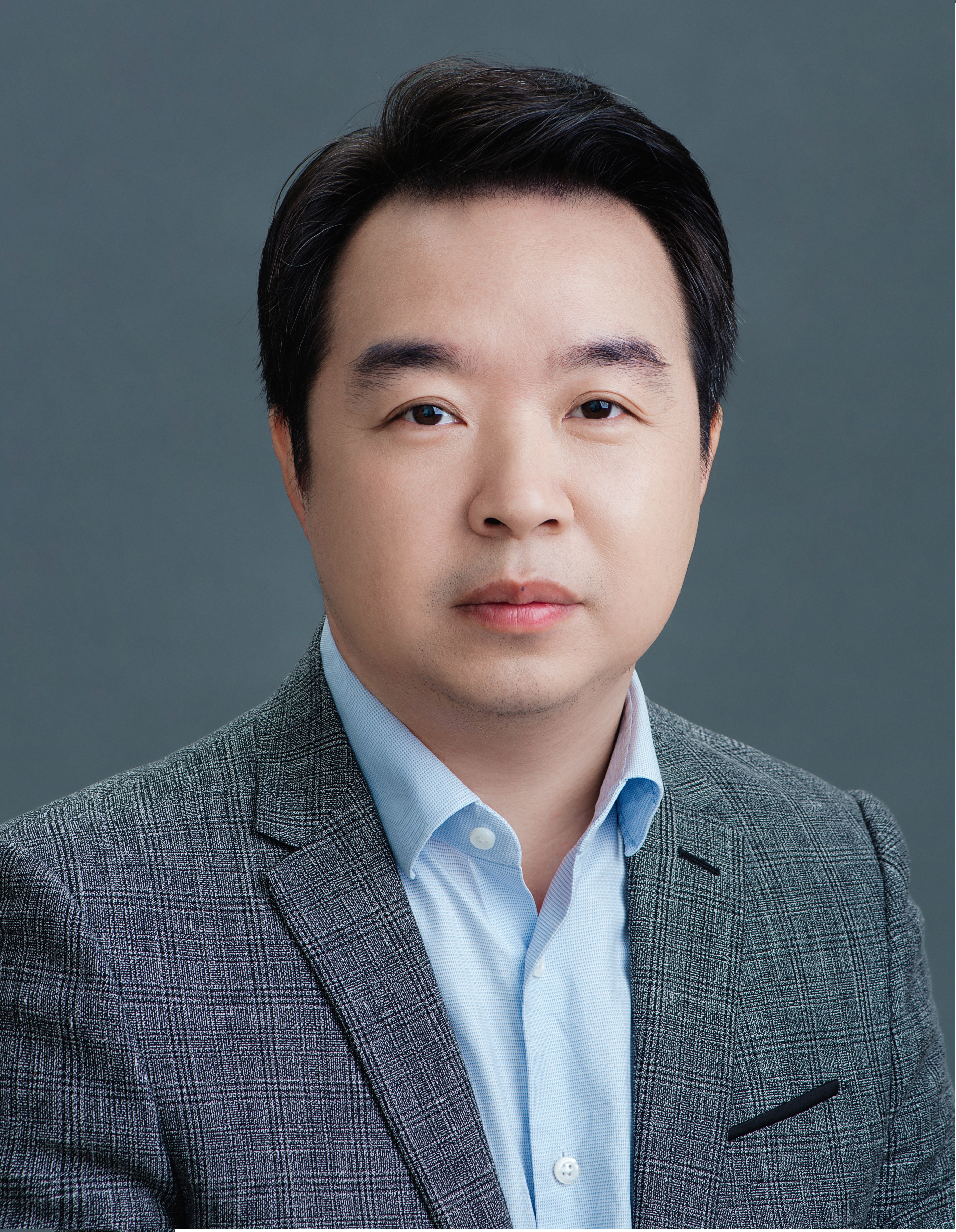}}]{Qinghua Hu}
(SM’13) received the B.S., M.S., and Ph.D. degrees from the Harbin Institute of Technology, Harbin, China, in 1999, 2002, and 2008, respectively. After that he joined Department of Computing, The Hong Kong Polytechnical University as a postdoctoral fellow. He became a full professor with Tianjin University in 2012, and now is a Chair Professor and Deputy Dean at College of Intelligence and Computing. His research interest is focused on uncertainty modeling, multi-modality learning, incremental learning and continual learning these years, funded by National Natural Science Foundation of China and The National Key Research and Development Program of China. He has published more than 300 peer-reviewed papers in IEEE TKDE, IEEE TPAMI, IEEE TNNLS, etc. He was a recipient of the best paper award of ICMLC 2015 and ICME 2021. He is an Associate Editor of the IEEE Transactions on Fuzzy Systems, ACTA AUTOMATICA SINICA, and ACTA ELECTRONICA SINICA.
\end{IEEEbiography}



\end{document}